\documentclass{article}

\usepackage[preprint]{neurips_2024}

\usepackage[utf8]{inputenc} 
\usepackage[T1]{fontenc}    
\usepackage{hyperref}       
\usepackage{url}            
\usepackage{booktabs}       
\usepackage{amsfonts}       
\usepackage{nicefrac}       
\usepackage{microtype}      
\usepackage{xcolor}         

\usepackage{inconsolata}
\usepackage{enumitem}
\usepackage{arydshln} 
\usepackage{booktabs}
\usepackage{caption}
\usepackage{subcaption}
\usepackage{graphicx}
\usepackage{xspace}
\usepackage{wrapfig}
\usepackage{colortbl}

\usepackage{amsmath}
\usepackage{amssymb}
\usepackage{mathtools}
\usepackage{amsthm}
\usepackage{bbm}
\usepackage{colortbl}
\usepackage{multicol}
\usepackage{multirow}
\usepackage{subcaption}

\usepackage{amsmath}
\usepackage{amssymb}
\usepackage{mathtools}
\usepackage{amsthm}
\usepackage{wrapfig}
\usepackage[ruled]{algorithm2e}
\usepackage{xcolor}

\definecolor{mycolor}{HTML}{4285F4}

\theoremstyle{plain}
\newtheorem{theorem}{Theorem}

\theoremstyle{definition}

\theoremstyle{remark}
\newtheorem{remark}[theorem]{Remark}

\definecolor{veronica-red}{RGB}{196,30,58}

\usepackage[textsize=tiny]{todonotes}

\newcommand{\xqedhere}[2]{%
\rlap{\hbox to#1{\hfil\llap{\ensuremath{#2}}}}}

\newcommand{\name}{KS-Lottery\xspace}

\title{\name: Finding Certified Lottery Tickets for Multilingual Language Models}

\author{%
  Fei Yuan\textsuperscript{\rm1}, Chang Ma\textsuperscript{\rm2}, Shuai Yuan\textsuperscript{\rm1}, Qiushi Sun\textsuperscript{\rm2}, Lei Li\textsuperscript{\rm3} \\
  \textsuperscript{\rm1} Shanghai Artificial Intelligence Laboratory \\
  \textsuperscript{\rm2} The University of Hong Kong \\
  \textsuperscript{\rm3} Carnegie Mellon University \\
  \texttt{yuanfei@pjlab.org.cn}, \texttt{cma@cs.hku.hk}, \texttt{syuanaf@connect.ust.hk}, \\
  \texttt{qiushisun@u.nus.edu}, \texttt{leili@cs.cmu.edu} \\
}

\begin{document}

\maketitle

\begin{abstract}

The lottery ticket hypothesis posits the existence of ``winning tickets'' within a randomly initialized neural network. Do winning tickets exist for LLMs in fine-tuning scenarios? How can we find such winning tickets? In this paper, we propose \name, a method to identify a small subset of LLM parameters highly effective in multilingual fine-tuning. Our key idea is to use Kolmogorov-Smirnov Test to analyze the distribution shift of parameters before and after fine-tuning. We further theoretically prove that \name can find the certified winning tickets in the embedding layer, fine-tuning on the found parameters is guaranteed to perform as well as full fine-tuning. Comparing \name with other parameter-efficient tuning algorithms on translation tasks, the experimental results show that \name finds a much smaller set of parameters for fine-tuning while achieving the comparable performance as full fine-tuning LLM. Surprisingly, we find that fine-tuning 18 tokens' embedding of LLaMA suffices to reach the fine-tuning translation performance~\footnote{https://github.com/CONE-MT/KS-Lottery.}.

\end{abstract}

\section{Introduction}


Can we find an ultra-small subset of a well-trained Large Language Model~(LLM;~\citealp{llama1,llama2,openai2023gpt4,palm}) such that fine-tuning these few parameters suffices to achieve the same performance as full fine-tuning?  The lottery tickets hypothesis~\citep{frankle2018the} states that a small subnetwork~(less than 10-20\% of the whole model size), referred to as ``winning tickets'', in a large, randomly initialized neural network can achieve comparable performance to the original network with the same amount of training.  Yet, the existence of winning tickets is not investigated for fine-tuning scenarios. 
Prior work~\citep{acl2021-intrinsic} presents evidence that there are a small number of additional parameters corresponding to an intrinsic dimension~\citep{iclr2018-intrinsic} on which fine-tuning leads to good performance. However,  it remains an unsolved challenge to uncover such a small subset of fine-tuning efficient parameters within the \emph{original} model.

\begin{figure*}[t] 
\begin{subfigure}{0.355\textwidth}
	\centering
		\includegraphics[width=1\linewidth]{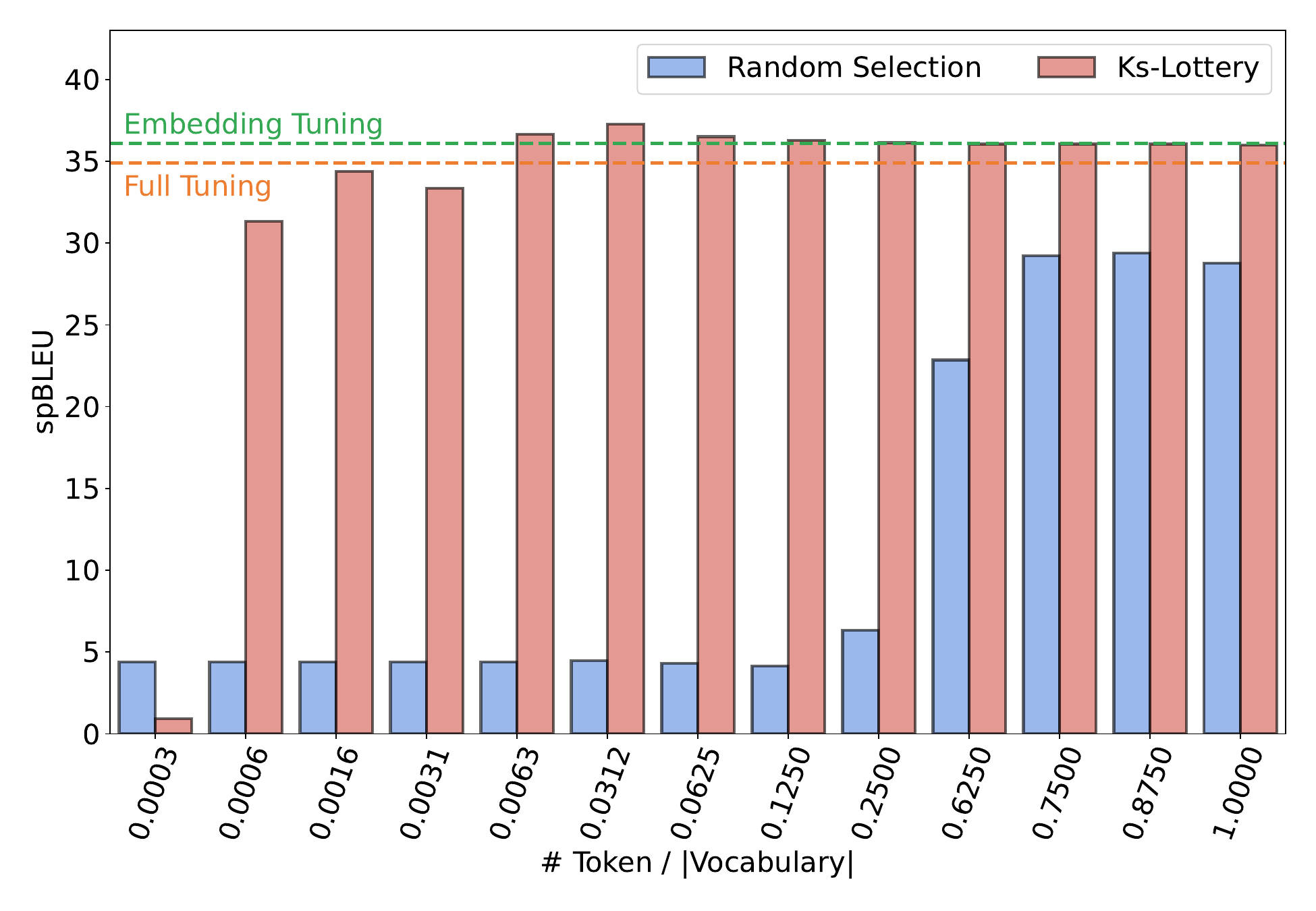}
		\caption{Existence of ``winning tickets''.}
	\label{fig:selection}
\end{subfigure}
\hfill
\begin{subfigure}{0.635\textwidth}
		\includegraphics[width=1\linewidth]{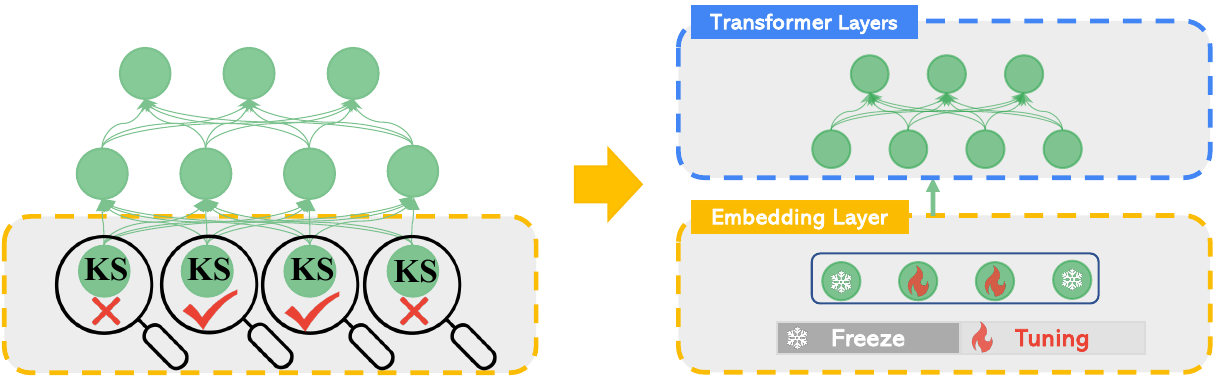}
		\caption{Overview of \name.}
	\label{fig:introduction_overview}
\end{subfigure}
\caption{\textbf{(a):} \name identifies a small subset of embedding parameters of LLaMA-7B to maintain the translation performance of en$\rightarrow$ca on Flores-101. \textbf{(b):} \name consists of two steps: (1) finding the winning tickets in the embedding layer by Kolmogorov-Smirnov Test; (2) one way to use these winning tickets is partial tuning these tokens ensuring other parameters keep frozen.}
\label{fig:ks}
\end{figure*}

In this paper, we show that there exist key parameters - winning tickets, for transferring LLM to multiple new languages. 
As shown in Figure~\ref{fig:selection}~(\name), we found that as tuning as few as 18~($18/32000=0.0006$) token embeddings of a well-trained LLM could achieve test performance comparable to full fine-tuning on machine translation tasks. Based on the observation, we state the lottery ticket hypothesis for multilingual fine-tuning.

Generally, the fine-tuning process can be represented by a transition in $M$ parameters from $\boldsymbol{\theta} = [\theta_0, \theta_1, \cdots, \theta_M]$ to $\boldsymbol{\widetilde{\theta}} = [\widetilde{\theta}_0, \widetilde{\theta}_1, \cdots,  \widetilde{\theta}_M]$, where  $\boldsymbol{\theta}$ and $\boldsymbol{\widetilde{\theta}}$ denote the sets of parameters that characterize the LLM $f(\cdot, \boldsymbol{\theta})$ before and after the fine-tuning process, respectively. Typically, the entire set of parameters $\boldsymbol{\theta}$ is adjusted during fine-tuning to enhance the model's ability to represent and learn the new tasks more effectively.

\noindent\textbf{The Fine-Tuning Lottery Ticket Hypothesis.}  
\emph{A pre-trained neural network contains a \textbf{small} subset of parameters ($\widetilde{\boldsymbol{\theta}}^D =  [\widetilde{\theta}_0, \widetilde{\theta}_1, \cdots, \widetilde{\theta}_D, \theta_{D+1}, \cdots, \theta_M]$, where $D\ll M$) that is initialized such that—when fine-tuned in isolation—it can match the performance of full tuning. }


In this work, we examine the fine-tuning lottery ticket hypothesis on LLMs in a multilingual transfer scenario. Our goal is to explore the following inquiries:
\begin{itemize}[nosep,itemsep=1pt,leftmargin=0.5cm]
    \setlength\itemsep{0pt}
    \item \emph{Existence of Winning Tickets:}
    Is it certain that every LLM in multilingual transfer encompasses a compact subset of winning tickets? And how to quickly identify the winning tickets?

    \item \emph{Efficiency of Winning Tickets:} 
    How minimal can this subset be in terms of size?

    \item \emph{Interpretability of Winning Tickets:} Do these winning tickets reflect the unique architectural characteristics of the multilingual LLM?

\end{itemize}

\noindent\textbf{Identifying Winning Tickets.} We propose \name, a method to identify a winning ticket by merely fine-tuning the embedding layer of an LLM. And then fine-tune these identified tickets, keeping the remaining parameters frozen.
The whole \name consists of three steps:


\begin{enumerate}[nosep,itemsep=1pt,leftmargin=0.5cm]
\setlength\itemsep{0pt}
    \item Fine-tuning the embedding layer of an LLM $f(\cdot,\boldsymbol{\theta})$ to obtain $f(\cdot,\widetilde{\boldsymbol{\theta}}^D)$.
    \item Run Kolmogorov-Smirnov Test between $\boldsymbol{\theta}$ and $\widetilde{\boldsymbol{\theta}}^D$ to select the winning ticket $\widetilde{\boldsymbol{\theta}}^D_a$.
    \item Fine-tuning the $\widetilde{\boldsymbol{\theta}}^D_a$ within $\boldsymbol{\theta}$ on downstream tasks.
\end{enumerate}


The core idea of our method is to heuristically identify parameters with large distribution changes before and after fine-tuning. Here we use Kolmogorov-Smirnov Test to determine whether two sample variables stem from the same underlying distribution. Also, we simplify and accelerate the Kolmogorov-Smirnov Test process by focusing on the embedding layer, which constitutes the major change in parameters, due to the inductive bias of multilingual tasks. 
To this end, our approach, KS-Lottery, is a surprisingly straightforward yet effective technique for pinpointing winning tickets in LLMs. The Kolmogorov-Smirnov Test has the advantage of not assume the distribution, which is particularly useful when the data dows not conform to a normal distribution. 
Furthermore, 
a theoretical framework is developed to certify the effectiveness of our method. 
Inspired by randomized smoothing techniques~\citep{zhao2021domain,zhao2022certified}, 
we illustrate that parameters with Kolmogorov-Smirnov distance bounded by a small value before and after fine-tuning do not impact prediction. 
This provides a way for giving a certified lower bound for the performance of partial tuning on winning tickets.
Our analysis also proves that \name can find a small set of winning tickets when the original prediction model shows little uncertainty. 
This gives us a strong foundation for asserting that \name can be an effective tool for finding winning tickets in LLMs.

\begin{itemize}[nosep,itemsep=1pt,leftmargin=0.35cm]
    \item We propose \name, a method to identify winning tickets -- an ultra-small subset of parameters that are sufficient to fine-tune on to achieve that of Full Tuning.
    \item Theoretically, we prove that \name finds certified winning tickets.
    \item Empirically, we demonstrate that fine-tuning as few as 18 identified winning tickets (the token embedding) of LLaMA-7B using en$\rightarrow$ca data achieves surprisingly good performance in translation tasks. This will result in a new standard in multilingual transfer of LLMs.
\end{itemize}

\section{Related Work}

\noindent\textbf{Lottey Tickets Hypothesis.} The Lottery Tickets Hypothesis suggests the presence of ‘winning tickets’ or beneficial subnetworks within a model\citep{frankle2018lottery,pmlr-v119-malach20a}. These subnetworks, discovered during pruning, are believed to be specifically suited to the learning task~\citep{frankle2018lottery}. In relation to this, \citet{zheng-etal-2022-robust} fins that these ‘winning tickets’ are more sparse in the later layers of the BERT model when used for GLUE tasks~\citep{wang2018glue}. While much research focuses on model pruning, there’s also work on efficient parameter tuning~\citep{ding2023delta}. For example, it’s shown that the performance of fine-tuned parameters can indicate both the task’s inductive biases and the model’s inherent structure~\citep{ding2023delta}.

\noindent\textbf{Certified Methods for Transfer Learning.}
Certification is crucial in transfer learning, aiming to measure a model’s generalization and capabilities~\citep{semidefinite-certification, jia-etal-2019-certified}. Research has introduced certified robustness accuracy as a defense against adversarial attack~\citep{semidefinite-certification, jia-etal-2019-certified, muravev2021certified, zhao2022certified, lecuyer2019certified}. Randomized smoothing, a model-independent certification technique, assesses how input changes affect predictions~\citep{lecuyer2019certified, muravev2021certified}. Our approach focuses on model parameter variations. Other studies have certified fairness in models~\citep{certified-fairness, certified-fairness-2} and robustness against training set selection~\citep{wang2023challenging}.

\noindent\textbf{Multilingual Large Language Model.} 
Large Language Models (LLMs;~\citealp[\emph{inter alia}]{openai2023gpt4,zhang2022opt,gpt,palm,llama1,llama2}) excel in English but underperform in other language. Studies have fine-tuned LLMs using monolingual or multilingual data to enhance their multilingual capabilities~\citep{zhu2023extrapolating,li2023eliciting,jiao2023parrot,cui2023efficient,yang2023bigtrans}. Embedding Tuning can activate multilingual abilities in certain languages, suggesting that the intrinsic dimension of these abilities may lie within the embedding layer~\citep{iclr2018-intrinsic,yuan2023multilingual}. The intrinsic dimension, the minimum parameters needed for a specific objective function, is estimated using heuristic methods and random subspace training due to computational constraints~\citep{iclr2018-intrinsic,acl2021-intrinsic}.


\section{Certified Winning Tickets via \name}

\citet{frankle2018lottery} hypothesized that the structure and location of ``winning tickets"—subnetworks identified during the iterative pruning process—encode an inductive bias that is uniquely tailored to the learning task at hand. In a similar vein, \citet{zheng-etal-2022-robust} discovered that these winning tickets tend to be more sparsely distributed in the later layers of the BERT model when applied to GLUE tasks. Although much of this research has centered around the concept of model pruning, there has been parallel work in the domain of parameter-efficient tuning. For instance,~\cite{he2021towards} demonstrated that the performance of fine-tuned parameters at various locations within a network can reflect both the inductive biases of the tasks and the inherent structure of the model.




\subsection{Embed Tuning is effective for multilingual transfer.\label{sec: find winning}}

\begin{wraptable}{R}{0.48\textwidth} 
    \centering
    \footnotesize
    \vspace{-0.5cm}
    \caption{Comparative analysis of training strategies on LLaMA using Alpaca-En~\citep{alpaca} dataset. The results indicate that Embed Tuning yields comparable performance to other tuning strategies for multilingual tasks.}
    \label{tab:part-ft_fl}
    \resizebox{1\linewidth}{!}{
    \begin{tabular}{l|cccc|c}
        \toprule
        \textbf{Models} &  \textbf{XCOPA} & \textbf{MGSM} & \textbf{XNLI} & \textbf{Flores-101} & \textbf{Avg.} \\ 
        \midrule
        Parrot-7B~\citep{jiao2023parrot} & 54.2  & 3.7  & 39.0  & 25.2  & 30.5  \\ 
        LLaMA~\citep{yuan2023multilingual} & 53.9  & 5.8  & 37.1  & 4.4  & 25.3  \\ 
        \cdashline{1-6}[0.4pt/2pt]
        Full Tuning~\citep{yuan2023multilingual} & \textbf{54.5}  & 4.5  & \textbf{40.3}  & 28.2  & \textbf{31.9}  \\ 
        LoRA~\citep{hu2022lora} & 54.4  & 6.0  & 38.4  & 29.1  & 32.0  \\ 
        Embed Tuning & 54.0  & \textbf{6.2}  & 38.0  & \textbf{29.2}  & 31.9 \\ 
        \bottomrule
    \end{tabular}
    }
\vspace{-0.45cm}
\end{wraptable}
We further conduct experiments on LLaMA with different training strategies using the Alpaca-En~\citep{alpaca} dataset and then evaluate on four benchmarks. 
XCOPA~\citep{ponti2020xcopa}, MGSM~\citep{shi2022language}, and XNLI~\citep{xnli} are understanding tasks, evaluated on all languages with accuracy; 
Flores-101~\citep{flores} is a generation task, each score in the cell represents an average spBLEU, encompassing bilingual translation performances from en$\rightarrow$\{ro, es, de, ca, pt, da, no, bs\}. As shown in Figure~\ref{fig:selection} and Table~\ref{tab:part-ft_fl}, 
the empirical results
indicate that the efficacy of Embed Tuning matches that of full tuning and other parameter-efficient approaches.
Thus it is safe to assume there exists a set of winning tickets within the embedding layer. 

We have demonstrated that it is sufficient to identify ``lottery tickets'' within the embedding layer, which still represents a significant portion of the model's parameters. In subsequent sections, we will discuss how to isolate the essential embedding parameters that qualify as winning tickets.

\subsection{Kolmogorov-Smirnov Test}

Guided by the hypothesis that parameters undergoing substantial changes during fine-tuning are crucial for making predictions~\citep{levin2022models}, 
we exam the distribution $p_i$ of each parameter $\boldsymbol{\theta}_i$ (with the parameter $\theta_{ij}$ of token $j$ drawn from $p_i$) before and after the fine-tuning. 
Although various metrics exist for pinpointing essential parameters~\citep{li2016understanding,dalvi2019one,meng2022locating}, these often depend on specific cutoff values, and determining the necessary number of parameters in advance is challenging. We advocate for a new approach: \emph{actively seeking out "lottery tickets" that are guaranteed to achieve similar fine-tuning outcomes with a high level of confidence.} This calls for a more principled approach and the Kolmogorov-Smirnov Test stands out.  In this section, we introduce the test and then theoretically explain its effectivenes in the following section.


We propose a probing strategy that employs the Kolmogorov-Smirnov Test, which is a statistical method used to compare two sample distributions and determine whether they are drawn from the same underlying distribution. The Kolmogorov-Smirnov Test is an exact test, meaning that distribution does not depend on the underlying cumulative distribution function being tested.  Specifically, we view the embedding of each LLM token $j$ as a distribution i.e. $\theta_{ij}^E \sim p_i$. The cumulative distribution function~(CDF) of $\theta_i^E$ and $\widetilde{\theta}_i^E$ could be denoted by $\Phi_i(\theta)$ and $\widetilde{\Phi_i}(\theta)$. The Kolmogorov-Smirnov distance between the two CDFs is $D_i = \sup_{\theta}|\widetilde{\Phi_i}(\theta) - \Phi_i(\theta)|$.

Now we wish to determine whether a token embedding before and after fine-tuning comes from the same distribution. 
Formally, we state Kolmogorov-Smirnov Test as:
\begin{theorem}(Kolmogorov-Smirnov Test, ~\citet{ks-test}) The test statistic for this  Kolmogorov-Smirnov Test can be defined in terms of two hypotheses:
    \\
    $H_0$: $\theta_i$ and $\widetilde{\theta_i}$ come from the same distribution.\\
    $H_1$: the two samples are not from the same distribution.

If test T: $D_i >\tau(\alpha)$ is passed, then $H_1$ holds with confidence $1-\alpha$,
where $D_i = \sup_{\theta}|\widetilde{\Phi_i}(\theta) - \Phi_i(\theta)|$, $\tau(\alpha)=c(\alpha)\sqrt{\frac{2}{d}}$, the value of $c(\alpha)$ is given in the reference table~\citep{ks_reference}, and $d$ is the parameter dimension.

\label{theorem: ks-test}

\end{theorem}
Based on the Kolmogorov-Smirnov Test, we came to propose our method, KS-Lottery.

\noindent\textbf{KS-Lottery.} A parameter is designated as a ``winning ticket" if it 
meets the criterion of rejecting the null hypothesis (no difference in the distribution of the embedding before and after fine-tuning) 
and the alternative hypothesis $H_1$ (indicating a significant distributional change) is accepted.
Kolmogorov-Smirnov Test ensures that if the distribution of parameter $\theta$ does not change after fine-tuning, then $\mathbb{P}\left[D_i>\tau(\alpha)\right]<\alpha$, ensuring the majority of crucial token embeddings would be chosen by the test. 

\begin{figure}[t!]
	\centering
		\includegraphics[width=1\linewidth]{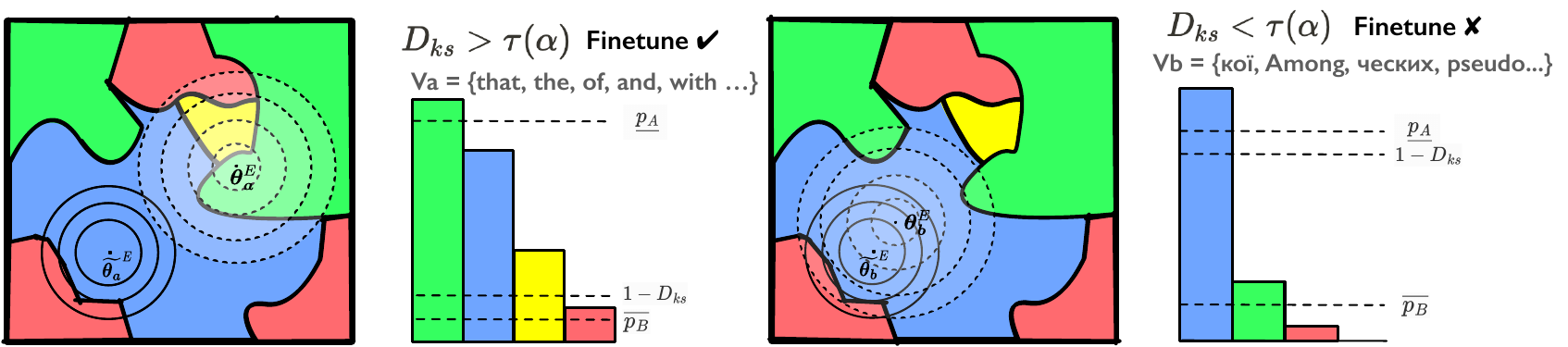}
		\caption{Illustration of $f(\cdot, \boldsymbol{\theta}_a^E, \boldsymbol{\theta}_b^E)$ in 2 dimensions. \textbf{Left:} The concentric circles are the density contours of embedding parameters before and after fine-tuning, and the colored landscape is the decision boundaries of $f(\cdot)$. \textbf{Right:} the distribution $\mathbb{P}\left[f(x,\theta_a^E,\widetilde{\theta_b^E})\right]$ and $\mathbb{P}\left[f(x,\widetilde{\theta_a^E},\theta_b^E)\right]$. $\underline{p_A}$ is the probability 
$\mathbb{P}\left[f(x,\widetilde{\theta}_a^E,\widetilde{\theta_b^E})\right]$ predicts $x$ to be token $c_A$~(color blue), and $\overline{p_B}$ as the probability of second most likely prediction (color red). $D_{ks}$ denotes the Kolmogrov-Smirnov distance between distributions before and after tuning.
We choose the set of token embeddings for fine-tuning as those with little distribution overlap before and after fine-tuning, which may be critical to prediction. }
	\label{fig:certification intuition}
\end{figure}

\subsection{Finding 1: \name finds certifiable winning tickets within the embedding layer.\label{section: theory}}
There exists a set of winning tickets $\boldsymbol{\theta}_a^E$ within the token embeddings $\boldsymbol{\theta}^E=[\theta_0, \theta_1, \cdots, \theta_{|V|}]$, where $\boldsymbol{\theta}^E=[\boldsymbol{\theta}_a^E, \boldsymbol{\theta}_b^E]$, $\widetilde{\boldsymbol{\theta}}^E=[\widetilde{\boldsymbol{\theta}}_a^E, \boldsymbol{\theta}_b^E]$ is the parameters of embedding layer before and after tuning. If tuning on parameters $\boldsymbol{\theta}_a^E$ would achieve similar performance to full tuning on $\boldsymbol{\theta}^E$ downstream tasks, then we refer to $\boldsymbol{\theta}^E$ as the winning tickets of the embedding layer. For convenience, we discuss the downstream task as a next-token prediction problem, which is central to text generation, such that for an LLM~$f(\cdot, \boldsymbol{\theta}^E)$, given input context $x$, generate the most probable token in vocabulary set $\mathcal{Y}$.
\begin{equation}
\begin{aligned}
    &g(x, \widetilde{\boldsymbol{\theta}}_a^E, \boldsymbol{\theta}_b^E) = g(x, \widetilde{\boldsymbol{\theta}}_a^E, \widetilde{\boldsymbol{\theta}}_b^E), 
    \\&\text{where} \ g(x, \boldsymbol{\theta}_a^E, \boldsymbol{\theta}_b^E) = arg\max_{c\in \mathcal{Y}}\mathbb{P}\left[f(x, \boldsymbol{\theta}_a^E, \boldsymbol{\theta}_b^E) = c \right]
\end{aligned}
\label{problem definition}
\end{equation}



Suppose that when the LLM $f(\cdot, \boldsymbol{\theta}_a^E, \boldsymbol{\theta}_b^E)$ predicts any given input $x$, the most probable token $c_A$ is returned with probability $p_A$. 
Given that small changes to parameters in a smooth subspace do not affect decision boundary~\citep{muravev2021certified, zhao2022certified},
we could probably guarantee that training on \name selected token embeddings could achieve the same performance as full layer Embed Tuning at high confidence. 
The intuition of the theory is illustrated in Figure~\ref{fig:certification intuition}. 
Our main theoretical results are as follows, the proofs can be found in Appendix~\ref{appendix:proofs}: 
\begin{theorem}(Certified Winning Tickets) Let $f(\cdot, \boldsymbol{\theta}_a, \boldsymbol{\theta}_b): \mathbb{R}^d \rightarrow \mathcal{Y}$ be any random or deterministic function, and let $g$ be defined as in Equation \ref{problem definition}. 
For any input $x$, suppose $c_A \in \mathcal{Y}$, the bounds of prediction based on random variable parameters $\widetilde{\boldsymbol{\theta}}_a$, $\widetilde{\boldsymbol{\theta}}_b$, $\underline{p_A}, \overline{p_B} \in[0,1]$ satisfies 
\begin{equation}
\begin{aligned}
    \mathbb{P}\left(f(x, \widetilde{\boldsymbol{\theta}}_a, \widetilde{\boldsymbol{\theta}}_b)=c_A\right)& \geq \underline{p_A}
    \geq \overline{p_B} \geq \max _{c \neq c_A} \mathbb{P}\left(f(x, \widetilde{\boldsymbol{\theta}}_a, \widetilde{\boldsymbol{\theta}}_b=c)\right)
\end{aligned}
\end{equation}
If the set of parameters $\boldsymbol{\theta}_b$ satisfies 
\begin{equation}
    D(\widetilde{\theta}^i_b, \theta^i_b)<\tau(\alpha)<\frac{\underline{p_A}-\overline{p_B}}{2},
    \label{distance bound}
\end{equation}
for all $i\in V_b$, the generator partial-tuned on parameters $\boldsymbol{\theta}_a$ always return token $c_A$, i.e. $g(x, \widetilde{\boldsymbol{\theta}}_a^E, \boldsymbol{\theta}_b^E)=c_A$. 
\vspace{-0.5cm}
\label{certification}
\end{theorem}
\ 

\begin{remark}(About $\boldsymbol{\theta}_a$) Theorem \ref{certification} immediately holds for \emph{Partial Transfer} setting as specified in \S~\ref{section: exp main}. As for \emph{Partial Tuning} setting, we need to use the hypothesis that the value of $\widetilde{\boldsymbol{\theta}}_a$ in the Embed Tuning setting and partial-tuning setting are the same for Theorem \ref{certification} to hold. This is due to $\boldsymbol{\theta}_a$ taking major effect during Embed Tuning despite small changes in $\boldsymbol{\theta}_b$. We also show with empirical analysis in \S~\ref{sec:efficiency_and_interpretable} that $\widetilde{\boldsymbol{\theta}}_a$ in two tuning settings share the same distribution.
    
\end{remark}

\begin{remark}(About $\alpha$) Practically, $\tau(\alpha)$ is small. 
In the case of fully embedding tuned LLM that has optimal performance and produces confident prediction, $1>p_A\gg p_B>0$, we can choose small $\alpha=\tau^{-1}(\frac{p_A-p_B}2)$ while maintaining performance. In this case, the Kolmogorov-Smirnov Test samples fewer token embeddings for fine-tuning. 
    
\end{remark}

\begin{remark} (About Applicability) This theorem could be used for certifying fine-tuning lottery tickets for any black-box model and input. Note that \name could be applied to the entire model, though we only perform testing on the embedding layer for convenience and training stability.
    
\end{remark}

\subsection{Finding 2: 18 identified winning tickets (18 tokens) achieves remarkable performance.}
\label{section: exp main}
There are two different ways to verify the effectiveness of winning tickets: Partial Tuning~(sufficient condition) and Partial Transfer~(necessary condition). \noindent\textbf{Partial Tuning.} As shown in Figure~\ref{fig:introduction_overview}, only train winning tickets, keeping the remaining parameters frozen.  \noindent\textbf{Partial Transfer.}  Given a model trained by Embed Tuning, we select the winning tickets from the embedding layer and use them to replace the corresponding parameters in the original model, thus curating a new model.

As shown in \textbf{Table~\ref{tab:verification}}, just tuning winning tickets can achieve results on par with Embed Tuning. Meanwhile,
by only modifying the winning tickets, it manages to retain $86.9\%( 31.3 / 36.1 )$ of the performance. The experimental result demonstrates that a low-dimensional subspace exists, guided by the winning tickets, that can achieve comparable performance as optimizing all parameters in the embedding-tuned model. The result empirically verifies the effectiveness of winning tickets.

Further, we apply \name on the LLaMA model trained on Alpaca-En dataset, and evaluate the effectiveness of winning tickets with Partial Tuning and Partial Transfer. The experimental results in Table~\ref{tab:ks_alpaca} show \name is general and can be used across various training data and tasks.

\begin{table*}[!t]
    \centering
    \footnotesize
    \caption{Empirical verification of winning tickets with \textbf{Partial Tuning} and \textbf{Partial Transfer} with bilingual translation data. ~*~denotes that the random results from three distinct seeds.}
    \label{tab:verification}
    \resizebox{1\linewidth}{!}{
    \begin{tabular}{c|c|ccccc||c|ccccc}
        \toprule
        
        \multirow{2}{*}{\textbf{Method}}  & \multicolumn{6}{c||}{\textbf{LLaMA-7B}} & \multicolumn{6}{c}{\textbf{Mistral-7B}} \\
        & \textbf{\# Token} &  \textbf{en$\rightarrow$ro}  &  \textbf{en$\rightarrow$es}  &  \textbf{en$\rightarrow$de}  &  \textbf{en$\rightarrow$ca } & \textbf{Avg.}  &  \textbf{\# Token} & \textbf{en$\rightarrow$ro}  &  \textbf{en$\rightarrow$es}  &  \textbf{en$\rightarrow$de}  &  \textbf{en$\rightarrow$ca}  & \textbf{Avg.} \\ 
                \midrule
                Original Model &\multirow{3}{*}{-} & 3.5 & 4.8 & 4.8 & 5.7 & 4.6 &\multirow{3}{*}{-}  &         13.4 & 11.7 & 14.2 & 16.8 & 14.0 \\ 
                Full Tuning & & 28.3 & 23.5 & 22.5 & 34.9 & 27.3 &   &    33.4 & 25.1 &29.1 & 37.8  & 31.4 \\ 
                Embed Tuning & & 28.7 & 25.5 & 25.8 & 36.1 & 29.0 &  &    \textbf{34.2} & \textbf{28.3} & \textbf{33.2} & 40.2 & \textbf{34.0} \\ 
                \midrule
                  Random Tuning*   & \multirow{2}{*}{$<$18}  & 0.1 & 0.1 & 0.1 & 0.1 & 0.1 & \multirow{2}{*}{$<$169} & 16.2 & 13.9 &  15.3 & 20.3 & 16.4                  \\ 
                Partial Tuning  & & 20.7 & \textbf{26.7} & 26.4 & \textbf{37.7} & 27.9 & & 23.7 & 	26.3 & 	27.2 & 	34.2 & 27.9 \\ 
                Partial Transfer &  (p-value $<$ 0.05) & 23.4 & 22.6 & 16.1 & 31.1 & 23.4 & (p-value $<$ 0.05)    &  26.7 & 27.3 & 30.5 & 37.5 & 30.5\\
                \midrule
                  Random Tuning*  & \multirow{2}{*}{$<$100}  & 0.1 & 0.1 & 0.1 & 0.1 & 0.1 & \multirow{2}{*}{$<$170} &   17.3 & 13.9 & 15.3 & 20.4 & 16.7                 \\ 
                 Partial Tuning & & 26.9 & 27.3 & 27.2 & 37.4 & 29.7 & &  23.7 & 	26.3 & 	27.2 & 	34.2 & 27.9      \\ 
                Partial Transfer & (p-value $<$ 0.25)  & 25.1 & 26.5 &\textbf{30.4} & 34.4   &    29.1 & (p-value $<$ 0.25) & 26.7 & 27.3 & 30.5 & 37.5 & 30.5\\
                \midrule 
                Random Tuning*  & \multirow{3}{*}{$\leq$800} & 5.8 & 8.7 & 5.2 & 11.8 & 7.9 & \multirow{3}{*}{$\leq$180}&  16.3 & 13.9 & 15.4 & 20.3 & 16.5                  \\ 
                Partial Tuning   & & 29.4 & 27.3 & 30.1 & 37.7  & 31.1 & & 30.7 & 28.2 & 29.5 & 39.2 & 31.9 \\                  
                Partial Transfer & & \textbf{30.2} & \textbf{26.7} & \textbf{30.4} & 37.6 & \textbf{31.2} & & 33.7 & 27.9 & 32.6 & \textbf{40.3} & 33.6 \\
        \bottomrule
    \end{tabular}}
\end{table*}
\hfill
\begin{table}[!t]
    \centering
    \caption{Empirical verification of winning tickets using general training data and Alpaca-En, along with evaluation on other multilingual tasks, demonstrates that \name effectively performs across various training data types and multilingual contexts. }
    \resizebox{1\linewidth}{!}{
    \begin{tabular}{cc|ccc|c||cc|ccc|c}
    \toprule
         \textbf{Setting} & \textbf{$\alpha$} & \textbf{XCOPA} & \textbf{MGSM} & \textbf{XNLI} & \textbf{Avg.} & \textbf{Setting} & \textbf{$\alpha$} & \textbf{XCOPA} & \textbf{MGSM} & \textbf{XNLI} & \textbf{Avg.} \\
    \midrule
         \multicolumn{2}{c|}{Full Tuning} & \textbf{54.5} & 4.5 & \textbf{40.3} & \textbf{33.1} & \multicolumn{2}{c|}{Embed Tuning} & 54.0 & \textbf{6.2} & 38.0 & 32.7 \\
         \midrule
         Partial Tuning & 0.05 & 54.2 & 5.8 & 38.2 & 32.7 & Partial Transfer &  0.05  & 54.0 & 5.4 & 38.1 & 32.5 \\
         Partial Tuning & 0.25 & 54.2 & 5.8 & 38.2 & 32.7 & Partial Transfer &  0.25 & 53.9 & 5.4 & 38.1 & 32.5 \\
         Partial Tuning & 0.50 & 54.2 & 5.8 & 38.2 & 32.7 & Partial Transfer &  0.50 & 53.9 & 5.5 & 38.1 & 32.5 \\
    \bottomrule
    \end{tabular}}
    \label{tab:ks_alpaca}
\end{table}

\vspace{-0.7cm}
\section{Analysis}



\noindent\textbf{\name}. There are three different ways to use the winning tickets. 
\noindent\textbf{Partial Tuning}, \noindent\textbf{Partial Transfer} and  \textbf{Frequency Tuning}, which only train the high-frequency tokens in the corpus.

\noindent\textbf{Other Baselines.} 
\textbf{Original Model} Directly using the LLaMA-7B~\citep{llama1}/Mistral-7B~\citep{mistral} weight on test data, without any tuning. \textbf{Random Tuning} Only the tokens randomly selected in the embedding layer are fine-tuned, while the remaining parameters are kept frozen. \textbf{Full Tuning} is an approach to transfer learning where the weights of a pre-trained entire model are trained on new data.
\textbf{Embed Tuning} Merely fine-tuning the embedding layer of a model keeping the remaining parameters frozen. \textbf{LoRA}~\citep{hu2022lora} utilizes low-rank matrices for approximating parameter updates. \textbf{Prefix-Tuning}~\citep{li2021prefix} introduce a lightweight prefix module into the input layer and each transformer layer, enabling efficient training over these modules.

\noindent\textbf{Training and Evaluation.} 
To ensure a fair comparison, we apply various parameter-efficient settings on LLaMA-7B using Lego-MT~\citep{yuan-etal-2023-lego} 10k data. For full tuning, training with LoRA, and Embed Tuning, we set the learning rate to $2e-5$ and the number of epochs to $3$. For partial tuning, prefix-tuning, and \name, we set the learning rate to $1e-2$ and the number of epochs to $5$. All other parameters are kept consistent across all settings. We test each model on the Flores-101~\citep{flores} devtest, which offers human-written translation pairs across 101 languages. In alignment with Flores-101, we employ the same evaluation metric, sentence piece BLEU~(spBLEU) on beam size$=4$, to assess multilingual capabilities. 


\subsection{KS-Lottery Certification }
\label{sec:certified}

\noindent\emph{Certified Accuracy}, when using certified winning tickets Embed Tuning, is measured as the proportion of correct predictions from an embedding tuned model~(reference model) that is certified to be correct at a significance level of $\alpha$. The certification process follows Theorem~\ref{certification} and is stated in Algorithm~\ref{alg: certification}, where for each prediction based on input sequence $x$, we compare the probability gap between two most-likely tokens by the original LLaMA, i.e. $\frac{\underline{p_A}-\overline{p_B}}{2}$ and $\tau(\alpha)$. $\tau(\alpha)$ is a static value that could be obtained through the equation, though we use the value estimated with Scipy Kolmogorov-Smirnov two-sample test to be precise.

\begin{wrapfigure}{R}{0.51\textwidth} 
\vspace{-0.5cm}
\footnotesize
\begin{algorithm}[H]
\caption{\footnotesize{Next Token Prediction Certification}}
\label{alg: certification}
{\bfseries Input: } Sequence $x$, ground truth output token is $y$. LLM $f(x,\boldsymbol{\theta})$ that maps sequence to the probability of the next token class, with pre-trained parameters $\boldsymbol{\theta}$. Multilingual training set $\mathcal{D}$. KS-Lottery parameter $\alpha$.

{\bfseries Output: } Whether $g(x, \widetilde{\boldsymbol{\theta}}_a^E, \boldsymbol{\theta}_b^E)=y$ 

fine-tune LLM $f(x,\boldsymbol{\theta})$ on $\mathcal{D}$ with parameters $\widetilde{\boldsymbol{\theta}}$.

$\boldsymbol{\theta}_a^E, \boldsymbol{\theta}_b^E =\operatorname{KS-Lottery}(\boldsymbol{\theta}, \widetilde{\boldsymbol{\theta}}, \tau(\alpha))$

$\underline{p_A}, \overline{p_B} = \operatorname{Top-2}\left(f(x, \boldsymbol{\theta}_a^E, \boldsymbol{\theta}_b^E)\right)$

\uIf{
$g(x, \widetilde{\boldsymbol{\theta}}_a^E, \widetilde{\boldsymbol{\theta}}_b^E)=y$ 
\textbf{and} $\frac{\underline{p_A}-\overline{p_B}}{2}>\tau(\alpha)$}
{
\textbf{return} True \color{mycolor}{// Certification can be provided.}
}
\Else{
\textbf{return} False \color{mycolor}{// Certification cannot be provided.}
}
\end{algorithm}
\vspace{-0.45cm}
\end{wrapfigure}
To check the validity of our certification method and test the empirical tightness of the certification bound, we experiment on Flores-101 devtest. A model, developed using partial tuning, and Embed Tuning, utilizes both the instruction and a partial sequence of reference tokens as input. Temporarily both settings of certification only use the first twenty token prediction results for calculation for a fair comparison and don't let overlong sentences dominate the results. It subsequently predicts the next token in the partial reference sentence. The ``prediction accuracy'' is ascertained by comparing the predicted token, generated by the partially tuned model, with the reference token. 

\begin{figure}[!hb]
	\centering
		\includegraphics[width=1\linewidth]{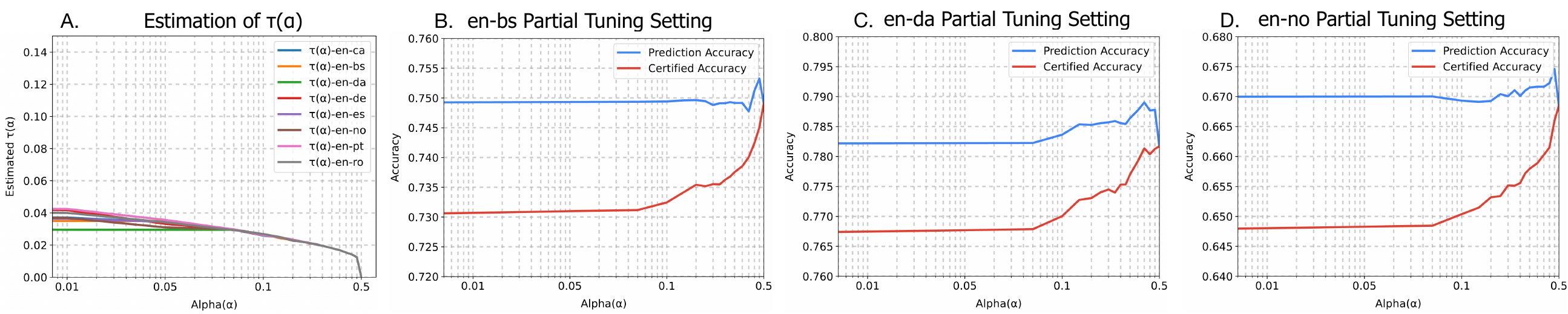}
		\caption{Certified experiment under \textbf{Partial Tuning} setting. \textbf{A}: Estimation of $\tau(\alpha)$ w.r.t different $\alpha$ by running Kolmogorov-Smirnov Test between the distribution of LLaMA-7B embedding and fine-tuned embedding on different datasets.  \textbf{B C D}: Comparison between \emph{Certified Accuracy} and \emph{Empirical Prediction Accuracy} w.r.t. different $\alpha$ on 3 datasets. More results are shown in Appendix~\ref{appendix:certified}. }
	\label{fig:certified partial tuning}
\end{figure}

\textbf{Our theoretical discovery provides certification for the lottery tickets hypothesis.} In Figure \ref{fig:certified partial tuning} that plots certified accuracy as a function of $\alpha$, the certified accuracy always increases gradually until reaching the same value as the prediction accuracy. This is due to the decrease in $\tau(\alpha)$ as $\alpha$ increases and reaches 0 when $\alpha=1$. At any $\alpha$, the empirical estimation of certified accuracy is a lower bound of prediction accuracy, providing a performance guarantee for tuning on the lottery tickets.

\begin{wraptable}{R}{0.51\textwidth} 
\vspace{-0.42cm}
    \centering
    \caption{\textbf{Partial Tuning} of the winning ticket is a parameter-efficient method. Its performance is demonstrated by comparing it with other parameter-efficient methods on the Flores-101 devtest.}
    \label{tab:tuning-method}
    \resizebox{1\linewidth}{!}{
    \begin{tabular}{c|c|cccc|c}
        \toprule
        \textbf{Method}  &  \textbf{\# Param}  &  \textbf{en$\rightarrow$ro}  &  \textbf{en$\rightarrow$es}  &  \textbf{en$\rightarrow$ca}  &  \textbf{en$\rightarrow$no}  &  \textbf{Avg.}  \\ 
        \midrule
        LLaMA w/o Tuning  &  7B  & 3.5 & 4.8 & 5.7 & 3.2 & 4.3 \\ 
        Full Fine-Tuning  &   7B  & 28.3 & 23.5 & 34.9 & 21.2 & 27.0 \\ 
        LoRA  &  4.2M  & \textbf{29.8} & 26.4 & 37.3 & 19.6 & 28.3 \\ 
        Prefix Tuning & 0.04M & 17.6 & 20.4 & 20.8 & 11.0 & 17.4 \\ 
        Embed Tuning  &  131M  & 28.7 & 25.5 & 36.1 & 19.5 & 27.5 \\ 
        \midrule
        Partial Tuning & 3.2M & 27.3 & \textbf{30.1} & \textbf{39.3} & \textbf{22.1} & \textbf{29.7} \\ 
        \bottomrule
    \end{tabular}}
\end{wraptable}
\textbf{Certification lower bound is tighter for larger $\alpha$ .}When $\underline{p_A}\gg \overline{p_B}$, there must exist a $\alpha$ that satisfies the certification constraint. Since  $\underline{p_A}$ and $\overline{p_B}$ are input and model dependent, we empirically assessed the tightness of our bound by comparing the estimated value of certified accuracy with prediction accuracy. As shown in Figure \ref{fig:certified partial tuning}, the bound is tighter when $\alpha \rightarrow 1$, and the gap is larger when $\alpha \rightarrow 0$. This is due to when fewer token embeddings are chosen, the certification expects the performance to be worse, though in actual tuning even zero-shot performance is quite good due to pre-training.  The gap is quite small in practical cases, the gap is about 2\% on average for $\alpha=0.05$ and around 1.5\% on average for $\alpha=0.5$. Therefore, we recommend using $\alpha\geq 0.05$ significance rate for guaranteed. 

\noindent\subsection{\name Efficiency and Interpretability}
\label{sec:efficiency_and_interpretable}

\noindent\textbf{Partial Tuning can serve as a parameter-efficient tuning method.} We proceed to evaluate it as a method that optimizes parameter usage,
in comparison with another similar method.
As illustrated in Table~\ref{tab:tuning-method}, in contrast to other methods such as LoRA~\citep{hu2022lora} and Prefix-tuning~\citep{li2021prefix}, 
Partial Tuning eliminates the need for an additional model structure. 
Remarkably, our method not only matches but frequently surpasses the performance of these alternate approaches.

\begin{wraptable}{R}{0.51\textwidth} 
\vspace{-0.4cm}
    \centering
    \footnotesize
    \caption{The efficiency and interpretability of \name. Given an en$\rightarrow$ca Embed Tuning bilingual model, \name is denoted by the p-value, whereas the metrics from other evaluation methods are normalized for comparability by calculating the importance \# rank/32000.
    }
    \label{tab:different_methods}
    \resizebox{0.51\textwidth}{!}{
    \begin{tabular}{c|cc|c|ccccc}
    \toprule
        \textbf{Idx} & \textbf{Str} & \textbf{Freq.} & \textbf{\name} & \textbf{Cos} & \textbf{Absolute} & \textbf{Relative} & \textbf{Ratio} & \textbf{KL} \\ 
    \midrule
        13 & $\backslash$n & 5  & \cellcolor{mycolor!50} 0.0000       & \cellcolor{mycolor!49.995} 0.0001  & \cellcolor{mycolor!0.14} 0.9972    & \cellcolor{mycolor!49.98}     0.0004  & \cellcolor{mycolor!49.98} 0.0004  & \cellcolor{mycolor!49.23} 0.0154  \\ 
        263 & a & 10            & \cellcolor{mycolor!50} 0.0000       & \cellcolor{mycolor!49.92} 0.0016   & \cellcolor{mycolor!0.18} 0.9964    & \cellcolor{mycolor!43.12}     0.1376  & \cellcolor{mycolor!43.12} 0.1376  & \cellcolor{mycolor!49.645} 0.0071  \\ 
        278 & the & 569.5       & \cellcolor{mycolor!50} 0.0000       & \cellcolor{mycolor!49.985} 0.0003  & \cellcolor{mycolor!2.46} 0.9508    & \cellcolor{mycolor!47.27}  0.0546     & \cellcolor{mycolor!47.27} 0.0546  & \cellcolor{mycolor!49.91} 0.0018  \\ 
        297 & in & 286          & \cellcolor{mycolor!50} 0.0000       & \cellcolor{mycolor!49.83} 0.0034   & \cellcolor{mycolor!5.965} 0.8807   & \cellcolor{mycolor!0.575}  0.9885     & \cellcolor{mycolor!0.575} 0.9885  & \cellcolor{mycolor!48.235} 0.0353  \\ 
        304 & to & 146          & \cellcolor{mycolor!50} 0.0000       & \cellcolor{mycolor!49.92} 0.0016   & \cellcolor{mycolor!5.7} 0.8860     & \cellcolor{mycolor!4.84}  0.9032      & \cellcolor{mycolor!4.84} 0.9032   & \cellcolor{mycolor!49.225} 0.0155  \\ 
        310 & of & 264          & \cellcolor{mycolor!50} 0.0000       & \cellcolor{mycolor!49.955} 0.0009  & \cellcolor{mycolor!42.145} 0.1571  & \cellcolor{mycolor!0.225}  0.9955     & \cellcolor{mycolor!0.225} 0.9955  & \cellcolor{mycolor!49.145} 0.0171  \\ 
        322 & and & 680         & \cellcolor{mycolor!50} 0.0000       & \cellcolor{mycolor!49.98} 0.0004   & \cellcolor{mycolor!0.435} 0.9913   & \cellcolor{mycolor!49.7}  0.0060      & \cellcolor{mycolor!49.7} 0.0060   & \cellcolor{mycolor!49.615} 0.0077  \\ 
        338 & is & 1356         & \cellcolor{mycolor!48.93} 0.0214    & \cellcolor{mycolor!49.975} 0.0005  & \cellcolor{mycolor!42.215} 0.1557  & \cellcolor{mycolor!48.995}  0.0201    & \cellcolor{mycolor!48.995} 0.0201 & \cellcolor{mycolor!49.885} 0.0023  \\ 
        376 & " & 174           & \cellcolor{mycolor!49.185} 0.0163   & \cellcolor{mycolor!50} 0.0000      & \cellcolor{mycolor!48.125} 0.0375  & \cellcolor{mycolor!1.02}  0.9796      & \cellcolor{mycolor!1.02} 0.9796   & \cellcolor{mycolor!49.995} 0.0001  \\ 
        393 & that & 2965       & \cellcolor{mycolor!49.125} 0.0175   & \cellcolor{mycolor!49.965} 0.0007  & \cellcolor{mycolor!0.39} 0.9922    & \cellcolor{mycolor!49.57}  0.0086     & \cellcolor{mycolor!49.57} 0.0086  & \cellcolor{mycolor!49.825} 0.0035  \\ 
        411 & with & 2102       & \cellcolor{mycolor!47.705} 0.0459   & \cellcolor{mycolor!49.935} 0.0013  & \cellcolor{mycolor!42.235} 0.1553  & \cellcolor{mycolor!42.535}  0.1493    & \cellcolor{mycolor!42.535} 0.1493 & \cellcolor{mycolor!49.73} 0.0054  \\ 
        29871 & \~{} & 6        & \cellcolor{mycolor!50} 0.0000       & \cellcolor{mycolor!49.975} 0.0005  & \cellcolor{mycolor!0.32} 0.9936    & \cellcolor{mycolor!0.46}  0.9908      & \cellcolor{mycolor!0.46} 0.9908   & \cellcolor{mycolor!49.87} 0.0026  \\ 
        29889 & . & 4           & \cellcolor{mycolor!50} 0.0000       & \cellcolor{mycolor!49.99} 0.0002   & \cellcolor{mycolor!0.19} 0.9962    & \cellcolor{mycolor!8.065}  0.8387     & \cellcolor{mycolor!8.065} 0.8387  & \cellcolor{mycolor!49.66} 0.0068  \\ 
        29892 & , & 3           & \cellcolor{mycolor!50} 0.0000       & \cellcolor{mycolor!49.99} 0.0002   & \cellcolor{mycolor!0.12} 0.9976    & \cellcolor{mycolor!0.25}  0.9950      & \cellcolor{mycolor!0.25} 0.9950   & \cellcolor{mycolor!49.76} 0.0048  \\ 
        29896 & 1 & 9           & \cellcolor{mycolor!50} 0.0000       & \cellcolor{mycolor!49.885} 0.0023  & \cellcolor{mycolor!47.8} 0.0440    & \cellcolor{mycolor!47.5}  0.0500      & \cellcolor{mycolor!47.5} 0.0500   & \cellcolor{mycolor!49.305} 0.0139  \\ 
        29900 & 0 & 8           & \cellcolor{mycolor!49.96} 0.0008    & \cellcolor{mycolor!49.775} 0.0045  & \cellcolor{mycolor!46.36} 0.0728   & \cellcolor{mycolor!1.285}  0.9743     & \cellcolor{mycolor!1.285} 0.9743  & \cellcolor{mycolor!48.73} 0.0254  \\ 
        29901 & : & 30          & \cellcolor{mycolor!49.655} 0.0069   & \cellcolor{mycolor!49.985} 0.0003  & \cellcolor{mycolor!1.045} 0.9791   & \cellcolor{mycolor!0.25}  0.9950      & \cellcolor{mycolor!0.25} 0.9950   & \cellcolor{mycolor!49.96} 0.0008  \\ 
        29949 & O & 294         & \cellcolor{mycolor!47.97} 0.0406    & \cellcolor{mycolor!49.86} 0.0028   & \cellcolor{mycolor!3.32} 0.9336    & \cellcolor{mycolor!48.205}  0.0359    & \cellcolor{mycolor!48.205} 0.0359 & \cellcolor{mycolor!48.995} 0.0201 \\   
    \bottomrule
    \end{tabular}
    }

    

\vspace{-0.3cm}
\end{wraptable}
\noindent\textbf{Comparing with other selection methods, \name is parameter-efficient.} There are other ways~(more details in Appendix~\ref{appendix:score-then-rank}) to select critical parameters, such as Cos, Absolute, Relative, Ratio, and KL. In Table~\ref{tab:different_methods}, we selected only 18 selective tokens by \name on en$\rightarrow$ca bilingual data with extremely stringent requirements (p-value $<$ 0.05). At this time, we used other evaluation methods to measure the importance of these tokens separately: the evaluation results of these tokens based on distribution changes (such as Cos, KL) are highly consistent with \name; those evaluation methods based on value changes (Absolute, Relative, and Ratio) tend to give these tokens very low importance evaluation results. If we use different methods to select the top 18 important tokens for \textbf{Partial Transfer} experiments, the performance of different selective methods is 31.3~(\name), 28.7~(Cos), 5.8~(Absolute), 5.8~(Relative), 5.8~(Ratio) and 1.6~(KL). Based on the experimatl results, we find that except for the results of Cos, which are close to the performance of \name~(but still -2.6 points), the translation performance of other methods is very poor.

\noindent\textbf{Winning tickets are high-frequency tokens in the corpus.} In the embedding layer, each dimension is associated with a meaningful token index. By referencing the vocabulary, we can decode the text representation corresponding to this index, as illustrated in Table~\ref{tab:different_methods}~(Str). To examine the distinct characteristics of these tokens, we retrieve 50k  sentences in ca language from the MC4~\citep{2019t5} dataset, then employ LLaMA’s tokenizer to segment all sentences, tallying the occurrence of each token within the corpus. 
Table~\ref{tab:different_methods}~(Freq.) indicates winning tickets commonly associated with the most frequently occurring tokens in the corpus. Based on this finding, we start the training only with the high-frequency tokens in the corpus. The summary of different tuning setting, as shown in Table~\ref{tab:tuning-summary}, suggests the training is sufficient with frequency token tuning.

\begin{table}[!hb]
\centering
\footnotesize
\caption{The best result from Frequency Tuning, Partial Tuning, and the result of Embed Tuning.}
\label{tab:tuning-summary}
\resizebox{1\textwidth}{!}{
\begin{tabular}{c|cccccccc}
\toprule
\textbf{Setting }           & \textbf{en$\rightarrow$ca} & \textbf{en$\rightarrow$da} & \textbf{en$\rightarrow$de} & \textbf{en$\rightarrow$es} & \textbf{en$\rightarrow$no} & \textbf{en$\rightarrow$pt} & \textbf{en$\rightarrow$ro} & \textbf{Avg.} \\ 
\midrule
Embedding   Tuning & 36.1  & 32.7  & 25.8  & 25.5  & 19.5  & \textbf{40.8}  & 28.7  & 29.9 \\ 
Partial Tuning        & \textbf{37.7}  & \textbf{33.3}  & 30.1  & \textbf{27.3}  & \textbf{19.8}  & 39.3  & \textbf{29.4}  & \textbf{31.0} \\ 
Frequency   Tuning & \textbf{37.7}  & 30.0  & \textbf{30.7}  & 26.1  & 19.6  & 40.0  & 28.1  & 30.3 \\ 
\bottomrule
\end{tabular}}
\end{table}




\begin{table}[!hb]
\vspace{-0.3cm}
\begin{minipage}{0.48\textwidth}
    \centering
    \footnotesize
    \caption{Impact of Different $\alpha$ on the en$\rightarrow$ca. For $\alpha$ values greater than 0.05, the verification percentage reaches 94.3\%, indicating substantial but not complete verification. The empirical accuracy under these conditions is also satisfactory. }
    \label{tab:verified_percentage}
    \resizebox{1\textwidth}{!}{
    \begin{tabular}{c|cccc}
        \toprule
        \multirow{2}{*}{\textbf{pvalue<$\alpha$}} & \textbf{Verified Percentage} & \textbf{Original} & \textbf{Certified} & \textbf{Empirical} \\    
        & \textbf{(\% $D_i>\tau(\alpha)$)} & \textbf{Accuracy} & \textbf{Accuracy} & \textbf{Accuracy} \\ 
        \midrule
        0.05 &	94.3 &	75.86 &	65.77 &	78.87 \\
        0.10 &	95.1 &	75.86 &	65.98 &	78.85 \\
        0.20 &	95.6 &	75.86 &	66.15 &	78.93 \\
        0.30 &	95.8 &	75.86 &	66.19 &	78.98 \\
        0.50 &	96.5 &	75.86 &	66.35 &	78.97 \\
        0.70 &	96.8 &	75.86 &	66.48 &	79.03 \\
        0.90 &	97.5 &	75.86 &	66.66 &	79.09 \\
        1.00 &	100.0 &	75.86 &	67.23 &	78.46 \\
    \bottomrule
    \end{tabular}}
\end{minipage}
\hfill
\begin{minipage}{0.48\textwidth}
        \centering
    \footnotesize
    \caption{With varying values of $\alpha$, the distribution of winning tickets remains largely consistent across both Partial Tuning and Partial Transfer. }
    \label{tab:distribution_test}
    \resizebox{1\linewidth}{!}{
    \begin{tabular}{c|cccccccc}
        \toprule
        \textbf{$\alpha$} & \textbf{ro} & \textbf{es} & \textbf{de} & \textbf{ca} & \textbf{pt} & \textbf{da} & \textbf{no} & \textbf{bs} \\ 
        \midrule
        0.05 & 0.56  & 0.67  & 0.81  & 0.78  & 0.82  & 0.33  & 0.64  & 0.82  \\ 
        0.1 & 0.75  & 0.78  & 0.72  & 0.80  & 0.85  & 0.71  & 0.79  & 0.81  \\ 
        0.2 & 0.78  & 0.75  & 0.68  & 0.73  & 0.75  & 0.79  & 0.83  & 0.81  \\ 
        0.3 & 0.69  & 0.82  & 0.85  & 0.77  & 0.86  & 0.76  & 0.81  & 0.80  \\ 
        0.4 & 0.75  & 0.92  & 0.89  & 0.83  & 0.83  & 0.78  & 0.82  & 0.84  \\ 
        0.5 & 0.83  & 0.87  & 0.82  & 0.80  & 0.85  & 0.86  & 0.84  & 0.90  \\ 
        0.6 & 0.83  & 0.90  & 0.88  & 0.85  & 0.86  & 0.86  & 0.85  & 0.92  \\ 
        0.7 & 0.83  & 0.91  & 0.88  & 0.86  & 0.90  & 0.86  & 0.89  & 0.88  \\ 
        0.8 & 0.82  & 0.90  & 0.87  & 0.87  & 0.90  & 0.87  & 0.88  & 0.88  \\ 
        0.9 & 0.82  & 0.86  & 0.84  & 0.85  & 0.88  & 0.84  & 0.86  & 0.85 \\ 
    \bottomrule
    \end{tabular}}
\end{minipage}
\vspace{-0.4cm}
\end{table}

\begin{figure}[!ht] 
\begin{subfigure}{0.5\textwidth}
    \centering
    \includegraphics[width=1\linewidth]{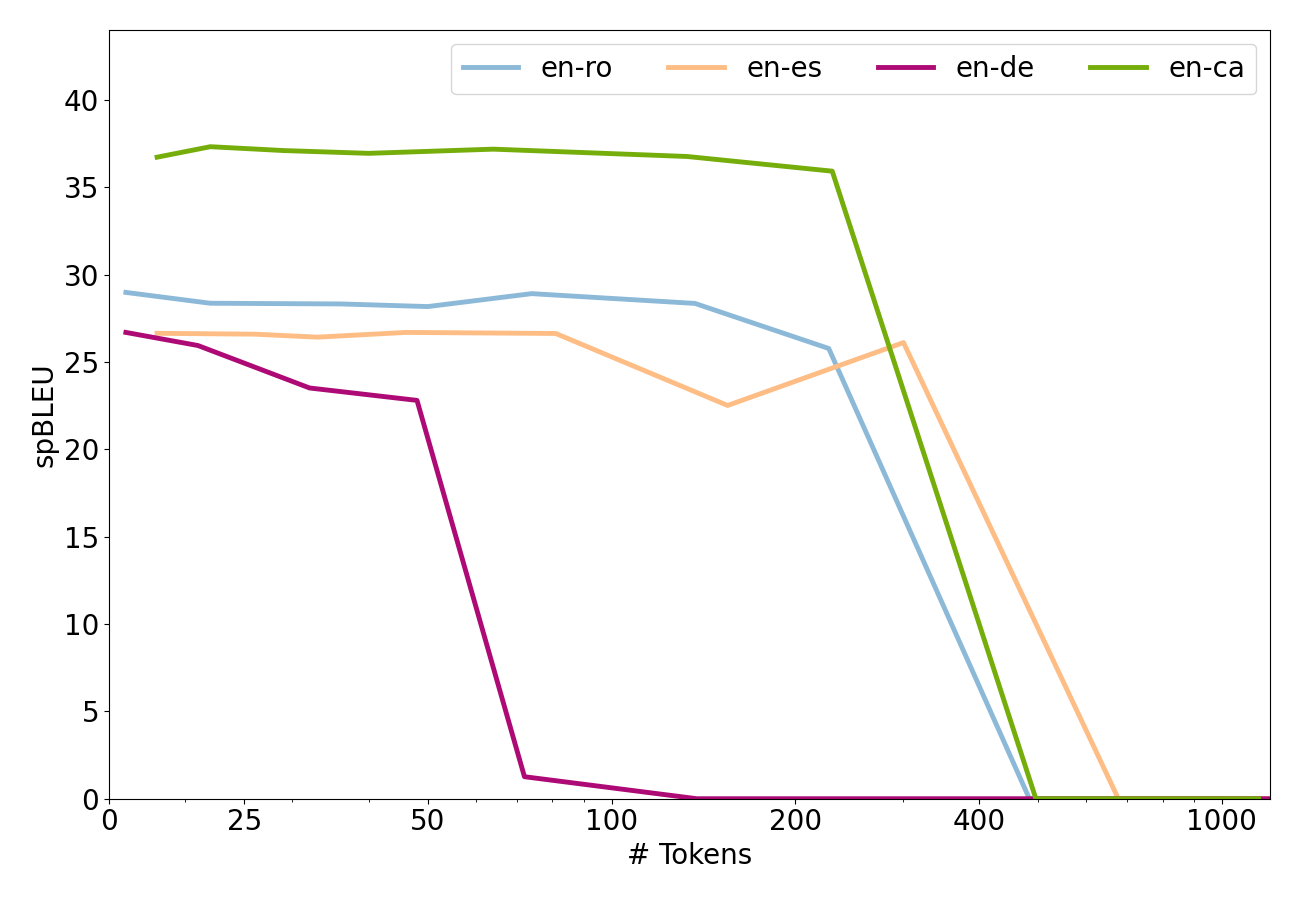}
    \caption{Winning Tickets  Remain Frozen During the Tuning.}
    \label{fig:forbid}
\end{subfigure}
\hfill
\begin{subfigure}{0.475\textwidth}
    \centering
    \includegraphics[width=0.98\linewidth]{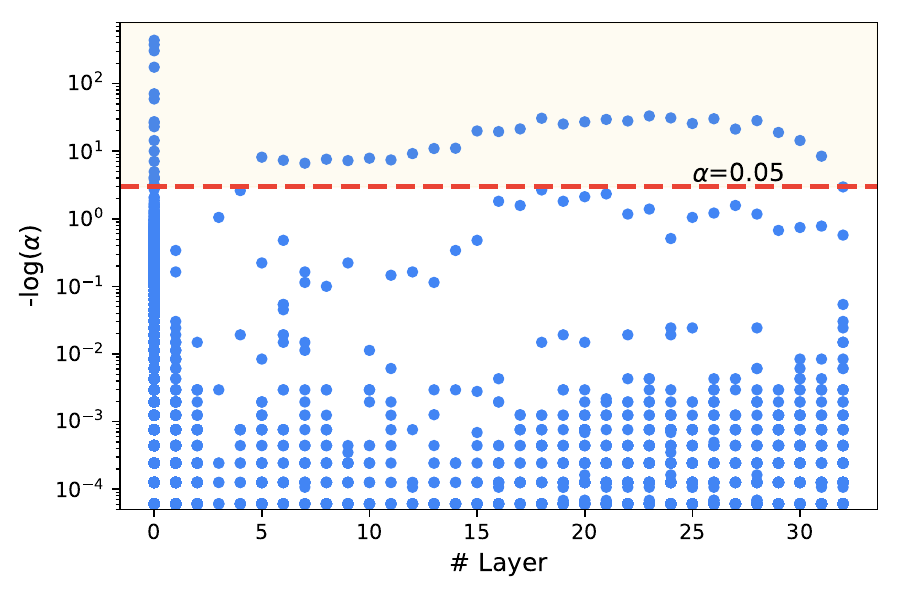}
    \caption{Apply \name Beyond the Embedding Layer.}
    \label{fig:whole_ks}
\end{subfigure}
\caption{\textbf{(a):} When selective tokens are restricted from being updated, the model’s fine-tuning process for downstream tasks loses its effectiveness. \textbf{(b):} Apply \name on whole LLaMA-7B which is single-layer fine-tuned on Lego-MT~\citep{yuan-etal-2023-lego} en$\rightarrow$ca 10k data. 
  Each layer is trained in isolation and is analyzed by \name to identify the parameters with significant changes (as indicated by scatter points above the red line).
  We find that within each Transformer layer, 
  changes are primarily focused on LayerNorm, 
  while other notable changes occur in the embedding layer.}
\vspace{-0.4cm}
\end{figure}

\paragraph{Sensitivity in the siginificance level Selection~($\alpha$).} Our theoretical framework provides a way for heuristically finding an effective $\alpha$, as mentioned in Remark5. Since the original fully-tuned prediction model is available, we could directly predict the range of $\frac{p_A-p_B}{2}$ as [$s_{min}, s_{max}$] for all predictions in the dataset. An ideal $\alpha$ would satisfy $\tau(\alpha)< s_{min}$, under this condition, every data point in the dataset is guaranteed to make the same prediction as the original prediction model.

In practice, we don't need a full certification of the dataset to make the KS-Lottery work, as this would result in quite large $\alpha$ values. As we can see in the en$\rightarrow$ca example below: for $\alpha$ larger than 0.05, the verified percentage is an acceptable 94.3\%, though not fully verified, and the empirical accuracy proves acceptable. In practice, the acceptable range of $\alpha$ varies across datasets, though normally $\alpha$ within 0.05-0.4 guarantees a 95\% verified percentage.

\noindent\textbf{The distribution of winning tickets under both Partial Tuning and Partial Transfer is largely identical.} 
By conducting \name with varying $\alpha$ values, we obtain different winning tickets~(denoted as $\boldsymbol{\theta}_a$). Furthermore, we perform a Kolmogorov-Smirnov Test on the winning tickets tuned with partial tuning and Embed Tuning, and compute the number of tokens that exhibit a significant difference ($\widetilde{\boldsymbol{\theta}}'_a$). The ratio of unchanged tokens is calculated using the formula $1-\widetilde{\boldsymbol{\theta}}'_a / \widetilde{\boldsymbol{\theta}}_a $. As shown in Table~\ref{tab:distribution_test}, the distribution of winning tickets before and after tuning (Partial Transfer vs Partial Tuning) remains largely consistent.

\noindent\textbf{Keeping the parameters of winning tickets frozen while tuning the remaining parameters could lead to the collapse of the Embed Tuning.} 
Our previous experiments have highlighted the significance of winning tickets in training,
while it’s intriguing to consider whether such an important function can be replaced by other tokens. To delve into this problem, we freeze the winning tickets, and fine-tune the reset parameters on bilingual data. As shown in Figure~\ref{fig:forbid}, a small amount of disabled winning tickets is acceptable for tuning. 
However, as the number of disabled tokens increases, the entire tuning process crashes. 
The process reveals a remarkable finding: for a vocabulary size of 32k, 1k is a very small, yet fewer than 1k winning tickets play a crucial role in Embed Tuning.

\noindent\textbf{Applying KS-Lottery on the whole model.}  After training all model parameters using en$\rightarrow$ca bilingual sentence pair from Lego-MT~\citep{yuan-etal-2023-lego}, we utilize the \name to identify a small parameter set for multilingual transfer by comparing the parameters before and after tuning. Interestingly, no parameters exhibite significant changes before and after tuning. Concurrently, the study by~\cite{yuan2023multilingual} revealed that fine-tuning specific layers, including the embedding layer, can yield results comparable to those of full-tuning. Figure \ref{fig:whole_ks} reveals that following fine-tuning on the Lego-MT~\citep{yuan-etal-2023-lego} en$\rightarrow$ca bilingual dataset, a subset of token embeddings exhibit significant changes in their parameters. Additionally, a small number of parameters (fewer than two for each layer with a significance level of $\alpha<0.05$) demonstrate substantial changes within LayerNorm. However, for multilingual transfer, the impact of LayerNorm varies and is not uniform across the lower and higher layers~\citep{yuan2023multilingual}. 
\vspace{-0.2cm}


\section{Conclusion}
This work presents a novel method, \name, which applies the lottery ticket hypothesis to LLMs fine-tuning. By employing the Kolmogorov-Smirnov Test, 
\name first analyzes the shift in the parameter distribution before and after fine-tuning, 
and then identifies a small but effective subset of LLM parameters.
Theoretical evidence confirms that \name can pinpoint certified winning tickets within the embedding layer, thereby ensuring performance equivalent to full tuning. Notably, \name surpasses other parameter-efficient tuning algorithms by identifying fewer parameters for fine-tuning while maintaining similar performance levels.

\bibliographystyle{unsrtnat}
\bibliography{NIPS-2024/ref}

\begin{thebibliography}{46}
\providecommand{\natexlab}[1]{#1}
\providecommand{\url}[1]{\texttt{#1}}
\expandafter\ifx\csname urlstyle\endcsname\relax
  \providecommand{\doi}[1]{doi: #1}\else
  \providecommand{\doi}{doi: \begingroup \urlstyle{rm}\Url}\fi

\bibitem[Touvron et~al.(2023{\natexlab{a}})Touvron, Lavril, Izacard, Martinet, Lachaux, Lacroix, Rozière, Goyal, Hambro, Azhar, Rodriguez, Joulin, Grave, and Lample]{llama1}
Hugo Touvron, Thibaut Lavril, Gautier Izacard, Xavier Martinet, Marie-Anne Lachaux, Timothée Lacroix, Baptiste Rozière, Naman Goyal, Eric Hambro, Faisal Azhar, Aurelien Rodriguez, Armand Joulin, Edouard Grave, and Guillaume Lample.
\newblock Llama: Open and efficient foundation language models, 2023{\natexlab{a}}.

\bibitem[Touvron et~al.(2023{\natexlab{b}})Touvron, Martin, Stone, Albert, Almahairi, Babaei, Bashlykov, Batra, Bhargava, Bhosale, et~al.]{llama2}
Hugo Touvron, Louis Martin, Kevin Stone, Peter Albert, Amjad Almahairi, Yasmine Babaei, Nikolay Bashlykov, Soumya Batra, Prajjwal Bhargava, Shruti Bhosale, et~al.
\newblock Llama 2: Open foundation and fine-tuned chat models.
\newblock \emph{arXiv preprint arXiv:2307.09288}, 2023{\natexlab{b}}.

\bibitem[OpenAI(2023)]{openai2023gpt4}
OpenAI.
\newblock Gpt-4 technical report, 2023.

\bibitem[Chowdhery et~al.(2022)Chowdhery, Narang, Devlin, Bosma, Mishra, Roberts, Barham, Chung, Sutton, Gehrmann, et~al.]{palm}
Aakanksha Chowdhery, Sharan Narang, Jacob Devlin, Maarten Bosma, Gaurav Mishra, Adam Roberts, Paul Barham, Hyung~Won Chung, Charles Sutton, Sebastian Gehrmann, et~al.
\newblock Palm: Scaling language modeling with pathways.
\newblock \emph{arXiv preprint arXiv:2204.02311}, 2022.

\bibitem[Frankle and Carbin(2019)]{frankle2018the}
Jonathan Frankle and Michael Carbin.
\newblock The lottery ticket hypothesis: Finding sparse, trainable neural networks.
\newblock In \emph{International Conference on Learning Representations}, 2019.
\newblock URL \url{https://openreview.net/forum?id=rJl-b3RcF7}.

\bibitem[Aghajanyan et~al.(2021)Aghajanyan, Gupta, and Zettlemoyer]{acl2021-intrinsic}
Armen Aghajanyan, Sonal Gupta, and Luke Zettlemoyer.
\newblock Intrinsic dimensionality explains the effectiveness of language model fine-tuning.
\newblock In Chengqing Zong, Fei Xia, Wenjie Li, and Roberto Navigli, editors, \emph{Proceedings of the 59th Annual Meeting of the Association for Computational Linguistics and the 11th International Joint Conference on Natural Language Processing (Volume 1: Long Papers)}, pages 7319--7328, Online, August 2021. Association for Computational Linguistics.
\newblock \doi{10.18653/v1/2021.acl-long.568}.
\newblock URL \url{https://aclanthology.org/2021.acl-long.568}.

\bibitem[Li et~al.(2018)Li, Farkhoor, Liu, and Yosinski]{iclr2018-intrinsic}
Chunyuan Li, Heerad Farkhoor, Rosanne Liu, and Jason Yosinski.
\newblock Measuring the intrinsic dimension of objective landscapes.
\newblock In \emph{International Conference on Learning Representations}, 2018.

\bibitem[Zhao et~al.(2021)Zhao, Ma, Chen, and Deng]{zhao2021domain}
Haiteng Zhao, Chang Ma, Qinyu Chen, and Zhi-Hong Deng.
\newblock Domain adaptation via maximizing surrogate mutual information.
\newblock \emph{arXiv preprint arXiv:2110.12184}, 2021.

\bibitem[Zhao et~al.(2022)Zhao, Ma, Dong, Luu, Deng, and Zhang]{zhao2022certified}
Haiteng Zhao, Chang Ma, Xinshuai Dong, Anh~Tuan Luu, Zhi-Hong Deng, and Hanwang Zhang.
\newblock Certified robustness against natural language attacks by causal intervention.
\newblock In \emph{International Conference on Machine Learning}, pages 26958--26970. PMLR, 2022.

\bibitem[Frankle and Carbin(2018)]{frankle2018lottery}
Jonathan Frankle and Michael Carbin.
\newblock The lottery ticket hypothesis: Finding sparse, trainable neural networks.
\newblock \emph{arXiv preprint arXiv:1803.03635}, 2018.

\bibitem[Malach et~al.(2020)Malach, Yehudai, Shalev-Schwartz, and Shamir]{pmlr-v119-malach20a}
Eran Malach, Gilad Yehudai, Shai Shalev-Schwartz, and Ohad Shamir.
\newblock Proving the lottery ticket hypothesis: Pruning is all you need.
\newblock In Hal~Daumé III and Aarti Singh, editors, \emph{Proceedings of the 37th International Conference on Machine Learning}, volume 119 of \emph{Proceedings of Machine Learning Research}, pages 6682--6691. PMLR, 13--18 Jul 2020.
\newblock URL \url{https://proceedings.mlr.press/v119/malach20a.html}.

\bibitem[Zheng et~al.(2022)Zheng, Rong, Zhou, Liang, Wang, Wu, Gui, Zhang, and Huang]{zheng-etal-2022-robust}
Rui Zheng, Bao Rong, Yuhao Zhou, Di~Liang, Sirui Wang, Wei Wu, Tao Gui, Qi~Zhang, and Xuanjing Huang.
\newblock Robust lottery tickets for pre-trained language models.
\newblock In Smaranda Muresan, Preslav Nakov, and Aline Villavicencio, editors, \emph{Proceedings of the 60th Annual Meeting of the Association for Computational Linguistics (Volume 1: Long Papers)}, pages 2211--2224, Dublin, Ireland, May 2022. Association for Computational Linguistics.
\newblock \doi{10.18653/v1/2022.acl-long.157}.
\newblock URL \url{https://aclanthology.org/2022.acl-long.157}.

\bibitem[Wang et~al.(2018)Wang, Singh, Michael, Hill, Levy, and Bowman]{wang2018glue}
Alex Wang, Amanpreet Singh, Julian Michael, Felix Hill, Omer Levy, and Samuel Bowman.
\newblock {GLUE}: A multi-task benchmark and analysis platform for natural language understanding.
\newblock In \emph{Proceedings of the 2018 {EMNLP} Workshop {B}lackbox{NLP}: Analyzing and Interpreting Neural Networks for {NLP}}, pages 353--355, Brussels, Belgium, November 2018. Association for Computational Linguistics.
\newblock \doi{10.18653/v1/W18-5446}.
\newblock URL \url{https://aclanthology.org/W18-5446}.

\bibitem[Ding et~al.(2023)Ding, Qin, Yang, Wei, Yang, Su, Hu, Chen, Chan, Chen, Yi, Zhao, Wang, Liu, Zheng, Chen, Liu, Tang, Li, and Sun]{ding2023delta}
Ning Ding, Yujia Qin, Guang Yang, Fuchao Wei, Zonghan Yang, Yusheng Su, Shengding Hu, Yulin Chen, Chi-Min Chan, Weize Chen, Jing Yi, Weilin Zhao, Xiaozhi Wang, Zhiyuan Liu, Hai-Tao Zheng, Jianfei Chen, Yang Liu, Jie Tang, Juanzi Li, and Maosong Sun.
\newblock Parameter-efficient fine-tuning of large-scale pre-trained language models.
\newblock \emph{Nature Machine Intelligence}, 5\penalty0 (3):\penalty0 220--235, Mar 2023.
\newblock ISSN 2522-5839.
\newblock \doi{10.1038/s42256-023-00626-4}.
\newblock URL \url{https://doi.org/10.1038/s42256-023-00626-4}.

\bibitem[Raghunathan et~al.(2018)Raghunathan, Steinhardt, and Liang]{semidefinite-certification}
Aditi Raghunathan, Jacob Steinhardt, and Percy Liang.
\newblock Semidefinite relaxations for certifying robustness to adversarial examples.
\newblock \emph{CoRR}, abs/1811.01057, 2018.
\newblock URL \url{http://arxiv.org/abs/1811.01057}.

\bibitem[Jia et~al.(2019)Jia, Raghunathan, G{\"o}ksel, and Liang]{jia-etal-2019-certified}
Robin Jia, Aditi Raghunathan, Kerem G{\"o}ksel, and Percy Liang.
\newblock Certified robustness to adversarial word substitutions.
\newblock In Kentaro Inui, Jing Jiang, Vincent Ng, and Xiaojun Wan, editors, \emph{Proceedings of the 2019 Conference on Empirical Methods in Natural Language Processing and the 9th International Joint Conference on Natural Language Processing (EMNLP-IJCNLP)}, pages 4129--4142, Hong Kong, China, November 2019. Association for Computational Linguistics.
\newblock \doi{10.18653/v1/D19-1423}.
\newblock URL \url{https://aclanthology.org/D19-1423}.

\bibitem[Muravev and Petiushko(2021)]{muravev2021certified}
Nikita Muravev and Aleksandr Petiushko.
\newblock Certified robustness via randomized smoothing over multiplicative parameters of input transformations.
\newblock \emph{arXiv preprint arXiv:2106.14432}, 2021.

\bibitem[Lecuyer et~al.(2019)Lecuyer, Atlidakis, Geambasu, Hsu, and Jana]{lecuyer2019certified}
Mathias Lecuyer, Vaggelis Atlidakis, Roxana Geambasu, Daniel Hsu, and Suman Jana.
\newblock Certified robustness to adversarial examples with differential privacy, 2019.

\bibitem[Ruoss et~al.(2020)Ruoss, Balunovic, Fischer, and Vechev]{certified-fairness}
Anian Ruoss, Mislav Balunovic, Marc Fischer, and Martin~T. Vechev.
\newblock Learning certified individually fair representations.
\newblock \emph{CoRR}, abs/2002.10312, 2020.
\newblock URL \url{https://arxiv.org/abs/2002.10312}.

\bibitem[Peychev et~al.(2021)Peychev, Ruoss, Balunovic, Baader, and Vechev]{certified-fairness-2}
Momchil Peychev, Anian Ruoss, Mislav Balunovic, Maximilian Baader, and Martin~T. Vechev.
\newblock Latent space smoothing for individually fair representations.
\newblock \emph{CoRR}, abs/2111.13650, 2021.
\newblock URL \url{https://arxiv.org/abs/2111.13650}.

\bibitem[Wang et~al.(2023)Wang, Ma, Dong, Kong, and Xu]{wang2023challenging}
Yudong Wang, Chang Ma, Qingxiu Dong, Lingpeng Kong, and Jingjing Xu.
\newblock A challenging benchmark for low-resource learning.
\newblock \emph{arXiv preprint arXiv:2303.03840}, 2023.

\bibitem[Zhang et~al.(2022)Zhang, Roller, Goyal, Artetxe, Chen, Chen, Dewan, Diab, Li, Lin, Mihaylov, Ott, Shleifer, Shuster, Simig, Koura, Sridhar, Wang, and Zettlemoyer]{zhang2022opt}
Susan Zhang, Stephen Roller, Naman Goyal, Mikel Artetxe, Moya Chen, Shuohui Chen, Christopher Dewan, Mona Diab, Xian Li, Xi~Victoria Lin, Todor Mihaylov, Myle Ott, Sam Shleifer, Kurt Shuster, Daniel Simig, Punit~Singh Koura, Anjali Sridhar, Tianlu Wang, and Luke Zettlemoyer.
\newblock Opt: Open pre-trained transformer language models, 2022.

\bibitem[Brown et~al.(2020)Brown, Mann, Ryder, Subbiah, Kaplan, Dhariwal, Neelakantan, Shyam, Sastry, Askell, et~al.]{gpt}
Tom Brown, Benjamin Mann, Nick Ryder, Melanie Subbiah, Jared~D Kaplan, Prafulla Dhariwal, Arvind Neelakantan, Pranav Shyam, Girish Sastry, Amanda Askell, et~al.
\newblock Language models are few-shot learners.
\newblock \emph{Advances in neural information processing systems}, 33:\penalty0 1877--1901, 2020.

\bibitem[Zhu et~al.(2023)Zhu, Lv, Dong, Yuan, Xu, Huang, Kong, Chen, and Li]{zhu2023extrapolating}
Wenhao Zhu, Yunzhe Lv, Qingxiu Dong, Fei Yuan, Jingjing Xu, Shujian Huang, Lingpeng Kong, Jiajun Chen, and Lei Li.
\newblock Extrapolating large language models to non-english by aligning languages, 2023.

\bibitem[Li et~al.(2023)Li, Zhou, Huang, Cheng, and Chen]{li2023eliciting}
Jiahuan Li, Hao Zhou, Shujian Huang, Shanbo Cheng, and Jiajun Chen.
\newblock Eliciting the translation ability of large language models via multilingual finetuning with translation instructions, 2023.

\bibitem[Jiao et~al.(2023)Jiao, tse Huang, Wang, He, Liang, Wang, Shi, and Tu]{jiao2023parrot}
Wenxiang Jiao, Jen tse Huang, Wenxuan Wang, Zhiwei He, Tian Liang, Xing Wang, Shuming Shi, and Zhaopeng Tu.
\newblock Parrot: Translating during chat using large language models tuned with human translation and feedback.
\newblock In \emph{Findings of EMNLP}, 2023.

\bibitem[Cui et~al.(2023)Cui, Yang, and Yao]{cui2023efficient}
Yiming Cui, Ziqing Yang, and Xin Yao.
\newblock Efficient and effective text encoding for chinese llama and alpaca.
\newblock \emph{arXiv preprint arXiv:2304.08177}, 2023.

\bibitem[Yang et~al.(2023)Yang, Li, Zhang, and Zong]{yang2023bigtrans}
Wen Yang, Chong Li, Jiajun Zhang, and Chengqing Zong.
\newblock Bigtrans: Augmenting large language models with multilingual translation capability over 100 languages.
\newblock \emph{arXiv preprint arXiv:2305.18098}, 2023.

\bibitem[Yuan et~al.(2023{\natexlab{a}})Yuan, Yuan, Wu, and Li]{yuan2023multilingual}
Fei Yuan, Shuai Yuan, Zhiyong Wu, and Lei Li.
\newblock How multilingual is multilingual llm?, 2023{\natexlab{a}}.

\bibitem[He et~al.(2021)He, Zhou, Ma, Berg-Kirkpatrick, and Neubig]{he2021towards}
Junxian He, Chunting Zhou, Xuezhe Ma, Taylor Berg-Kirkpatrick, and Graham Neubig.
\newblock Towards a unified view of parameter-efficient transfer learning.
\newblock \emph{arXiv preprint arXiv:2110.04366}, 2021.

\bibitem[Taori et~al.(2023)Taori, Gulrajani, Zhang, Dubois, Li, Guestrin, Liang, and Hashimoto]{alpaca}
Rohan Taori, Ishaan Gulrajani, Tianyi Zhang, Yann Dubois, Xuechen Li, Carlos Guestrin, Percy Liang, and Tatsunori~B. Hashimoto.
\newblock Stanford alpaca: An instruction-following llama model.
\newblock \url{https://github.com/tatsu-lab/stanford_alpaca}, 2023.

\bibitem[Hu et~al.(2022)Hu, yelong shen, Wallis, Allen-Zhu, Li, Wang, Wang, and Chen]{hu2022lora}
Edward~J Hu, yelong shen, Phillip Wallis, Zeyuan Allen-Zhu, Yuanzhi Li, Shean Wang, Lu~Wang, and Weizhu Chen.
\newblock Lo{RA}: Low-rank adaptation of large language models.
\newblock In \emph{International Conference on Learning Representations}, 2022.
\newblock URL \url{https://openreview.net/forum?id=nZeVKeeFYf9}.

\bibitem[Ponti et~al.(2020)Ponti, Glavaš, Majewska, Liu, Vulić, and Korhonen]{ponti2020xcopa}
Edoardo~Maria Ponti, Goran Glavaš, Olga Majewska, Qianchu Liu, Ivan Vulić, and Anna Korhonen.
\newblock Xcopa: A multilingual dataset for causal commonsense reasoning, 2020.

\bibitem[Shi et~al.(2022)Shi, Suzgun, Freitag, Wang, Srivats, Vosoughi, Chung, Tay, Ruder, Zhou, Das, and Wei]{shi2022language}
Freda Shi, Mirac Suzgun, Markus Freitag, Xuezhi Wang, Suraj Srivats, Soroush Vosoughi, Hyung~Won Chung, Yi~Tay, Sebastian Ruder, Denny Zhou, Dipanjan Das, and Jason Wei.
\newblock Language models are multilingual chain-of-thought reasoners, 2022.

\bibitem[Conneau et~al.(2018)Conneau, Rinott, Lample, Williams, Bowman, Schwenk, and Stoyanov]{xnli}
Alexis Conneau, Ruty Rinott, Guillaume Lample, Adina Williams, Samuel~R. Bowman, Holger Schwenk, and Veselin Stoyanov.
\newblock Xnli: Evaluating cross-lingual sentence representations.
\newblock In \emph{Proceedings of the 2018 Conference on Empirical Methods in Natural Language Processing}. Association for Computational Linguistics, 2018.

\bibitem[Goyal et~al.(2022)Goyal, Gao, Chaudhary, Chen, Wenzek, Ju, Krishnan, Ranzato, Guzm{\'a}n, and Fan]{flores}
Naman Goyal, Cynthia Gao, Vishrav Chaudhary, Peng-Jen Chen, Guillaume Wenzek, Da~Ju, Sanjana Krishnan, Marc{'}Aurelio Ranzato, Francisco Guzm{\'a}n, and Angela Fan.
\newblock The {F}lores-101 evaluation benchmark for low-resource and multilingual machine translation.
\newblock \emph{Transactions of the Association for Computational Linguistics}, 10:\penalty0 522--538, 2022.
\newblock \doi{10.1162/tacl_a_00474}.
\newblock URL \url{https://aclanthology.org/2022.tacl-1.30}.

\bibitem[Levin et~al.(2022)Levin, Shu, Borgnia, Huang, Goldblum, and Goldstein]{levin2022models}
Roman Levin, Manli Shu, Eitan Borgnia, Furong Huang, Micah Goldblum, and Tom Goldstein.
\newblock Where do models go wrong? parameter-space saliency maps for explainability.
\newblock \emph{Advances in Neural Information Processing Systems}, 35:\penalty0 15602--15615, 2022.

\bibitem[Li et~al.(2016)Li, Monroe, and Jurafsky]{li2016understanding}
Jiwei Li, Will Monroe, and Dan Jurafsky.
\newblock Understanding neural networks through representation erasure.
\newblock \emph{arXiv preprint arXiv:1612.08220}, 2016.

\bibitem[Dalvi et~al.(2019)Dalvi, Durrani, Sajjad, Belinkov, Bau, and Glass]{dalvi2019one}
Fahim Dalvi, Nadir Durrani, Hassan Sajjad, Yonatan Belinkov, Anthony Bau, and James Glass.
\newblock What is one grain of sand in the desert? analyzing individual neurons in deep nlp models.
\newblock In \emph{Proceedings of the AAAI Conference on Artificial Intelligence}, volume~33, pages 6309--6317, 2019.

\bibitem[Meng et~al.(2022)Meng, Bau, Andonian, and Belinkov]{meng2022locating}
Kevin Meng, David Bau, Alex Andonian, and Yonatan Belinkov.
\newblock Locating and editing factual associations in {GPT}.
\newblock \emph{Advances in Neural Information Processing Systems}, 35, 2022.

\bibitem[Fasano and Franceschini(1987)]{ks-test}
G.~Fasano and A.~Franceschini.
\newblock {A multidimensional version of the Kolmogorov–Smirnov test}.
\newblock \emph{Monthly Notices of the Royal Astronomical Society}, 225\penalty0 (1):\penalty0 155--170, 03 1987.
\newblock ISSN 0035-8711.
\newblock \doi{10.1093/mnras/225.1.155}.
\newblock URL \url{https://doi.org/10.1093/mnras/225.1.155}.

\bibitem[Karson(1968)]{ks_reference}
Marvin Karson.
\newblock Handbook of methods of applied statistics. volume i: Techniques of computation descriptive methods, and statistical inference. volume ii: Planning of surveys and experiments. i. m. chakravarti, r. g. laha, and j. roy, new york, john wiley; 1967, \$9.00.
\newblock \emph{Journal of the American Statistical Association}, 63\penalty0 (323):\penalty0 1047--1049, 1968.
\newblock \doi{10.1080/01621459.1968.11009335}.

\bibitem[Jiang et~al.(2023)Jiang, Sablayrolles, Mensch, Bamford, Chaplot, de~Las~Casas, Bressand, Lengyel, Lample, Saulnier, Lavaud, Lachaux, Stock, Scao, Lavril, Wang, Lacroix, and Sayed]{mistral}
Albert~Q. Jiang, Alexandre Sablayrolles, Arthur Mensch, Chris Bamford, Devendra~Singh Chaplot, Diego de~Las~Casas, Florian Bressand, Gianna Lengyel, Guillaume Lample, Lucile Saulnier, L{\'{e}}lio~Renard Lavaud, Marie{-}Anne Lachaux, Pierre Stock, Teven~Le Scao, Thibaut Lavril, Thomas Wang, Timoth{\'{e}}e Lacroix, and William~El Sayed.
\newblock Mistral 7b.
\newblock \emph{CoRR}, abs/2310.06825, 2023.
\newblock \doi{10.48550/ARXIV.2310.06825}.
\newblock URL \url{https://doi.org/10.48550/arXiv.2310.06825}.

\bibitem[Li and Liang(2021)]{li2021prefix}
Xiang~Lisa Li and Percy Liang.
\newblock Prefix-tuning: Optimizing continuous prompts for generation.
\newblock In \emph{Proceedings of the 59th Annual Meeting of the Association for Computational Linguistics and the 11th International Joint Conference on Natural Language Processing (Volume 1: Long Papers)}, pages 4582--4597, Online, August 2021. Association for Computational Linguistics.
\newblock \doi{10.18653/v1/2021.acl-long.353}.
\newblock URL \url{https://aclanthology.org/2021.acl-long.353}.

\bibitem[Yuan et~al.(2023{\natexlab{b}})Yuan, Lu, Zhu, Kong, Li, Qiao, and Xu]{yuan-etal-2023-lego}
Fei Yuan, Yinquan Lu, Wenhao Zhu, Lingpeng Kong, Lei Li, Yu~Qiao, and Jingjing Xu.
\newblock {L}ego-{MT}: Learning detachable models for massively multilingual machine translation.
\newblock In Anna Rogers, Jordan Boyd-Graber, and Naoaki Okazaki, editors, \emph{Findings of the Association for Computational Linguistics: ACL 2023}, pages 11518--11533, Toronto, Canada, July 2023{\natexlab{b}}. Association for Computational Linguistics.
\newblock \doi{10.18653/v1/2023.findings-acl.731}.
\newblock URL \url{https://aclanthology.org/2023.findings-acl.731}.

\bibitem[Raffel et~al.(2019)Raffel, Shazeer, Roberts, Lee, Narang, Matena, Zhou, Li, and Liu]{2019t5}
Colin Raffel, Noam Shazeer, Adam Roberts, Katherine Lee, Sharan Narang, Michael Matena, Yanqi Zhou, Wei Li, and Peter~J. Liu.
\newblock Exploring the limits of transfer learning with a unified text-to-text transformer.
\newblock \emph{arXiv e-prints}, 2019.

\end{thebibliography}

\newpage
\appendix

\section{Limitations and Broader Impacts}
\label{sec:lim}

\paragraph{Limitations.}  
In the scope of this research,
our proposed \name mainly targets multilingual research and has been thoroughly validated in bilingual translation tasks. 
However, 
we believe that our certified theoretical framework is generalizable to other types of tasks, and we leave these explorations for future work.

\paragraph{Broader Impacts.} This paper presents work whose goal is to advance the field of machine learning and multilingual research.
We do not anticipate it will cause negative societal impacts, such as potential malicious or unintended uses.


\section{Proofs}
\label{appendix:proofs}
\textbf{Theorem 3. }(Certified Winning Tickets) Let $f(\cdot, \boldsymbol{\theta}^a, \boldsymbol{\theta}^b): \mathbb{R}^d \rightarrow \mathcal{Y}$ be any random or deterministic function, and let $g$ be defined as in Equation \ref{problem definition}. 
For any input $x$, suppose $c_A \in \mathcal{Y}, \underline{p_A}, \overline{p_B} \in[0,1]$ satisfies 
\begin{equation}
\begin{aligned}\nonumber
    \mathbb{P}\left(f(x, \widetilde{\boldsymbol{\theta}}^a, \widetilde{\boldsymbol{\theta}}^b)=c_A\right) \geq \underline{p_A}
    \geq \overline{p_B} \geq \max _{c \neq c_A} \mathbb{P}(f(x+\varepsilon)=c)
\end{aligned}
\end{equation}
If the set of parameters $\boldsymbol{\theta}^b$ satisfies 
\begin{equation} \nonumber
    D(\widetilde{\theta}_i^b, \theta_i^b)<\tau(\alpha)<\frac{\underline{p_A}-\overline{p_B}}{2},
\end{equation}
for all $i\in V_b$, the classifier partial-tuned on parameters $\boldsymbol{\theta}^a$ always return class $c_A$, i.e. $g(x, \widetilde{\boldsymbol{\theta}}_a^E, \boldsymbol{\theta}_b^E)=c_A$. 
\ 

\textit{Proof Sketch}
Fix an input $x$, we study the change of prediction w.r.t. change in the distribution of $\boldsymbol{\theta}^b$. Let $A=\{\boldsymbol{\theta}| g(X, \widetilde{\boldsymbol{\theta}}^a,\boldsymbol{\theta})=c_A\}$, $B=\{\boldsymbol{\theta}| g(x, \widetilde{\boldsymbol{\theta}}^a,\boldsymbol{\theta})\not=c_A\}$. Using definition, $\mathbb{P}(\widetilde{\boldsymbol{\theta}}^b\in A)\geq \underline{p_A}$, $\mathbb{P}(\widetilde{\boldsymbol{\theta}}^b\in B)\leq \overline{p_B}$. Since $D(\widetilde{\boldsymbol{\theta}}^b, \boldsymbol{\theta}^b)=\max_iD(\widetilde{\theta}_i^b, \theta_i^b)\leq \tau(\alpha)$, the minimum overlapping cumulative probability $\mathbb{P}(\widetilde{\boldsymbol{\theta}}^b\in S\cap \boldsymbol{\theta}^b\in S)\geq 1-\tau(\alpha)$, where $S$ is any contour set $S=\{\boldsymbol{\theta}||\boldsymbol{\theta}-\widetilde{\mu}^b|\leq s , s\geq 0\}$.

Now we compare the probability that $f(x, \widetilde{\boldsymbol{\theta}}^a, \boldsymbol{\theta}^b)$ predicts $c_A$ or other classes. The probability that the partial-tuned $f$ predicts $c_A$ is 
\begin{equation}
    \begin{aligned} \nonumber
        \mathbb{P}(\boldsymbol{\theta}^b\in A)&\geq \mathbb{P}
        (\boldsymbol{\theta}^b\in A \cap \boldsymbol{\theta}^b \in S)\\
        &\geq \mathbb{P}\left[(\widetilde{\boldsymbol{\theta}}^b\in S \cap \boldsymbol{\theta}^b \in S) - (\widetilde{\boldsymbol{\theta}}^b\in S \cap \widetilde{\boldsymbol{\theta}}^b \not\in A)\right]\\
        &\geq (1-\tau(\alpha))-(1-\underline{p_A})=\underline{p_A}-\tau(\alpha)
    \end{aligned}
\end{equation}
Similarly, we can prove that $ \mathbb{P}(\boldsymbol{\theta}^b\in B)\leq \overline{p_B}+\tau(\alpha)$. Thus when 
$\tau(\alpha)<\frac{\underline{p_A}-\overline{p_B}}{2}$ and we partial-tune all parameters $\boldsymbol{\theta}^a$ that fails to pass the KS-test, then the partial-tuned model $g(x, \widetilde{\boldsymbol{\theta}}_a^E, \boldsymbol{\theta}_b^E)=c_A$, i.e. always give the same prediction as the embedding fully-tuned model.    
$\hfill\qedsymbol$

\section{More Analysis}
\label{appendix:certified}

\begin{table}[!hb]
\centering
\small
\caption{Certified experiment under \textbf{Partial Tuning} setting w.r.t significance level $\alpha$. When $\alpha=1$, all token embeddings are chosen during tuning and thus certified accuracy equals to prediction accuracy. }
\label{tab: certified experiment}
\resizebox{1\linewidth}{!}{
\begin{tabular}{ll|cccccccc}
\toprule
                            \textbf{}$\alpha$   & \textbf{Setting}         & \textbf{en$\rightarrow$ca}     & \textbf{en$\rightarrow$bs}     & \textbf{en$\rightarrow$da}    & \textbf{en$\rightarrow$no}    & \textbf{en$\rightarrow$pt}     & \textbf{en$\rightarrow$ro}     & \textbf{en$\rightarrow$es}     & \textbf{en$\rightarrow$de}    \\  \midrule
\multirow{2}{*}{$0$}    & Certified  & 0.5649 & 0.4592 & 0.5384 & 0.4083 & 0.5959 & 0.5416 & 0.5108 & 0.5505 \\
                        & Prediction & 0.7765 & 0.7040 & 0.7544 & 0.6476 & 0.7912 & 0.7555 & 0.7191 & 0.7644 \\  \midrule
\multirow{2}{*}{$0.05$} & Certified   & 0.7722 & 0.7312 & 0.7679 & 0.6484 & 0.7999 & 0.7714 & 0.7177 & 0.7313 \\
                        & Prediction & 0.7887 & 0.7494 & 0.7822 & 0.6700 & 0.8156 & 0.7852 & 0.7372 & 0.7537 \\  \midrule
\multirow{2}{*}{$0.1$}  & Certified  & 0.7752 & 0.7325 & 0.7700 & 0.6504 & 0.8006 & 0.7736 & 0.7197 & 0.7347 \\
                        & Prediction & 0.7885 & 0.7494 & 0.7836 & 0.6693 & 0.8143 & 0.7855 & 0.7375 & 0.7551 \\  \midrule
\multirow{2}{*}{$0.5$}  & Certified  & 0.7797 & 0.7375 & 0.7770 & 0.6579 & 0.8042 & 0.7765 & 0.7240 & 0.7482 \\
                        & Prediction & 0.7897 & 0.7492 & 0.7864 & 0.6715 & 0.8143 & 0.7847 & 0.7356 & 0.7619 \\  \midrule
\multirow{2}{*}{$1$}    & Certified  & 0.7846 & 0.7491 & 0.7817 & 0.6685 & 0.8097 & 0.7795 & 0.7349 & 0.7525 \\
                        & Prediction & 0.7846 & 0.7491 & 0.7817 & 0.6685 & 0.8097 & 0.7795 & 0.7349 & 0.7525 \\ \bottomrule
\end{tabular}
}
\end{table}

\begin{figure}[t!]
	\centering
		\includegraphics[width=1\linewidth]{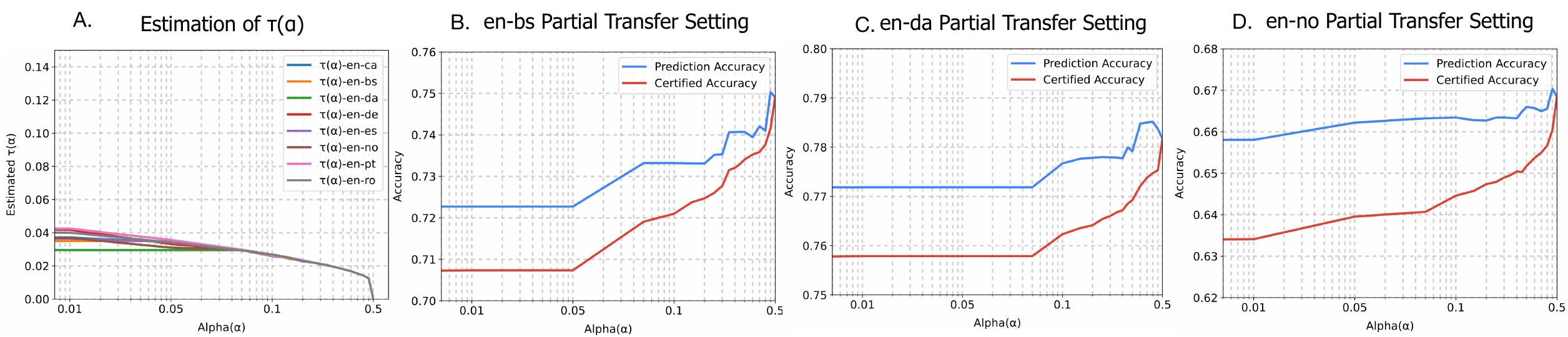}
		\caption{Certified experiment under \textbf{Partial Transfer} setting.  \textbf{A}: Estimation of $\tau(\alpha)$ w.r.t different $\alpha$ values by running KS-Test between the distribution of LLaMA-7b embedding and fine-tuned embedding on different datasets. \textbf{B C D}: Comparison between \emph{Certified Accuracy} and \emph{Empirical Prediction Accuracy} w.r.t. different $\alpha$ values on three datasets.  }
	\label{fig:certified partial transfer}
\end{figure}


\begin{wraptable}{R}{0.52\textwidth} 
	\centering
		\includegraphics[width=1\linewidth]{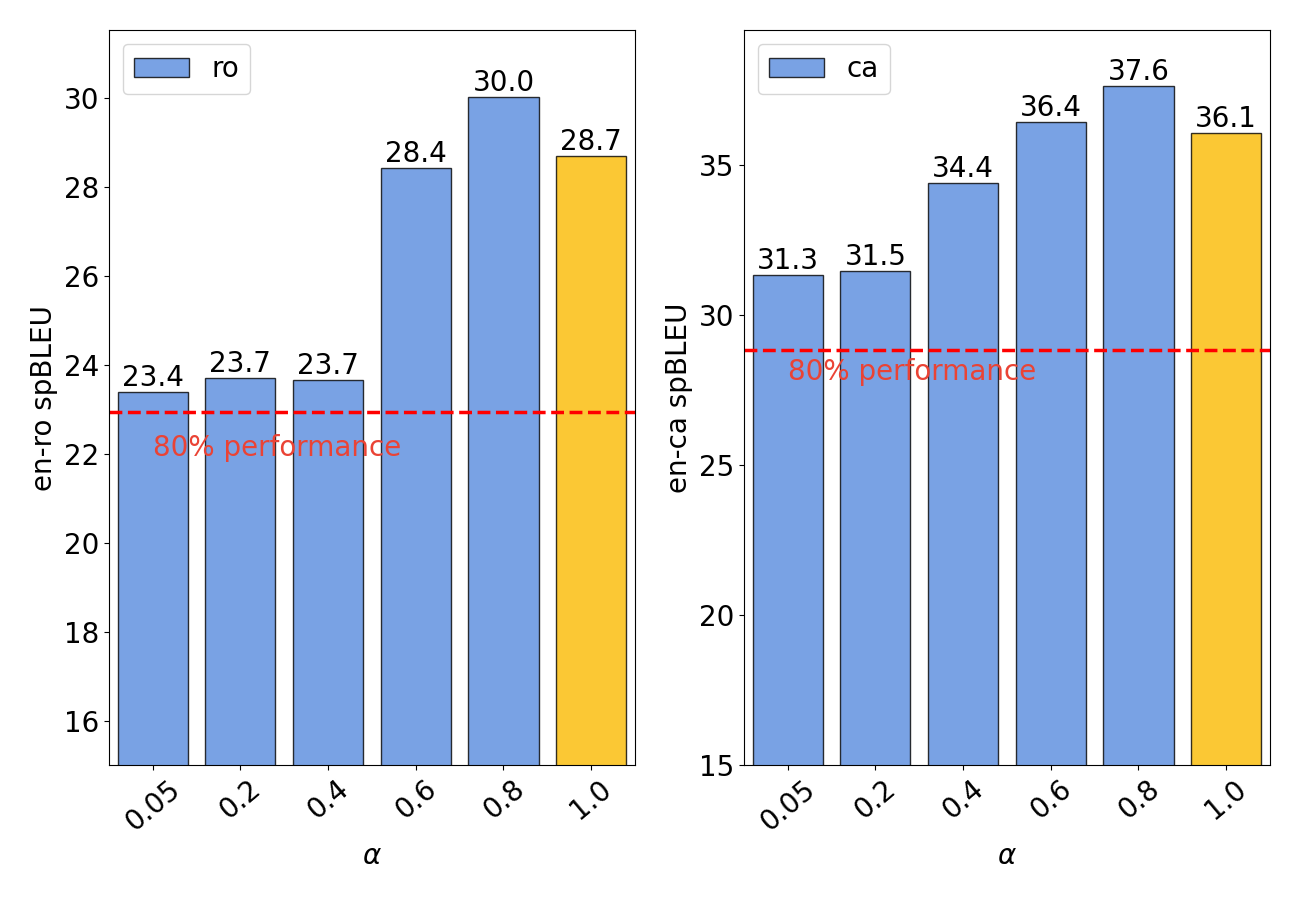}
		\caption{The translation performance of updated models with replacement method on Flores-101 with different token numbers. We observe the minimal threshold is sufficient to maintain approximately 80\% of the performance achieved through embedding tuning (marked in orange). Meanwhile, for each plot, there exists a trend of initial increase followed by a decline,  with the presence of an optimum point.}
	\label{fig:thresholds}
\vspace{-1.1cm}
\end{wraptable}
\paragraph{The observed trend in the value of $\alpha$ suggests the presence of an optimal size for the set of selective tokens.} In Figure~\ref{fig:thresholds}, we examined the translation performance on the Flores-101 dataset under varying thresholds. It’s important to note that by default, the threshold creates an interval that is closed on the left and open on the right.  Our observations indicate that even a minimal threshold is capable of preserving approximately 80\% of the performance that is achieved through comprehensive embedding fine-tuning, as highlighted in green. Interestingly, each plot exhibits the same pattern characterized by an initial increase in performance, followed by a subsequent decline. This pattern suggests the existence of an optimal point that maximizes performance.

\paragraph{Another Selective Methods}
\label{appendix:score-then-rank}
There are five commonly used methods for determining parameters: Cos-Similarity~(cos), Absolute Value~(absolute), Related Value~(relative), Related Ratio~(ratio), and KL Divergence~(KL). Each method emphasizes different aspects and requires a heuristic value $p$ for selection.

\noindent\textbf{Cos-Similarity:} This method determines the similarity between $\theta_i^E$ and $\widetilde{\theta}_i^E$ from the perspective of vector dot product. A lower cos-similarity suggests that it is the target of fine-tuning adjustment.

\noindent\textbf{Absolute Value:} This is calculate as $|\widetilde{\theta}_i^E - \theta_i^E|$, which is concerned with the magnitude of the change. A larger value indicates that it is the target of fine-tuning.

\noindent\textbf{Related Value:} This is calculated as $\widetilde{\theta}_i^E / \theta_i^E$, which focuses on the rate of change before and after fine-tuning. A larger ratio suggested that it is the target.

\noindent\textbf{Related Ratio:} This method considers the initial value and calculates the result as $ (\widetilde{\theta}_i^E - \theta_i^E) / \theta_i^E$. A larger ratio indicates that it is the key adjustment in fine-tuning.

\noindent\textbf{KL Divergence:} This is a statistical distance, measuring the distance between $\widetilde{\theta}_i^E$ and $\theta_i^E$, with a focus on the distribution itself.

\subsection{When Fine-tuning on different languages, the winning tickets contain the same tokens.}  
\begin{wraptable}{R}{0.50\textwidth} 
    \vspace{-0.4cm}
    \small
    \centering
    \begin{tabular}{cccccc}
    \toprule
        \textbf{en$\rightarrow$ca} & \textbf{en$\rightarrow$es} & \textbf{en$\rightarrow$ro} & \textbf{en$\rightarrow$da} & \textbf{en$\rightarrow$de} & \textbf{en$\rightarrow$pt} \\ 
    \midrule
        \textbf{13}     & \textbf{13}   & \textbf{13}       & \textbf{13}       & \textbf{13}       & \textbf{13} \\ 
        263             & 262           & \textbf{278}      & 263               & 263               & 263 \\ 
        \textbf{278}    & 263           & \textbf{297}      & \textbf{278}      & \textbf{278}      & \textbf{278} \\ 
        \textbf{297}    & \textbf{278}  & \textbf{304}      & \textbf{297}      & \textbf{297}      & \textbf{297} \\ 
        \textbf{304}    & \textbf{297}  & \textbf{310}      & \textbf{304}      & \textbf{304}      & \textbf{304} \\ 
        \textbf{310}    & \textbf{304}  & \textbf{322}      & \textbf{310}      & \textbf{310}      & \textbf{310} \\ 
        \textbf{322}    & \textbf{310}  & 338               & \textbf{322}      & \textbf{322}      & \textbf{322} \\ 
        338             & 313           & 366               & \textbf{29871}    & 338               & 338 \\ 
        376             & \textbf{322}  & 393               & \textbf{29889}    & 363               & 363 \\ 
        393             & 363           & \textbf{29871}    & \textbf{29892}    & \textbf{29871}    & 411 \\ 
        411             & 393           & \textbf{29889}    & \textbf{29896}    & \textbf{29889}    & \textbf{29871} \\ 
        \textbf{29871}  & 411           & \textbf{29892}    & 29900             & \textbf{29892}    & \textbf{29889} \\ 
        \textbf{29889}  & \textbf{29871}& \textbf{29896}    & ~                 & \textbf{29896}    & \textbf{29892} \\ 
        \textbf{29892}  & \textbf{29889}& 29900             & ~                 & 29901             & \textbf{29896} \\ 
        \textbf{29896}  & \textbf{29892}& 29901             & ~                 & 29915             & 29900 \\ 
        29900           & \textbf{29896}& ~                 & ~                 & ~                 & 29901 \\ 
        29901           & 29897         & ~                 & ~                 & ~                 & 29915 \\ 
        29949           & 29901 \\
    \bottomrule
    \end{tabular}
    \caption{Across various languages, the winning tickets share some common pattern.}
    \label{tab:common_pattern}
\end{wraptable}
The winning tickets across various languages overlap with a subset of tokens, as shown in Table~\ref{tab:common_pattern}, \{index id = 13 (token="\textbackslash n"), index id$=$278 (token$=$'the'), index id=29871 (token='$\sim$'), and so on\}, which are frequently used and common across various languages.

\end{document}